%% file: main.tex
\documentclass[10pt,journal,compsoc]{IEEEtran}
\ifCLASSOPTIONcompsoc
  \usepackage[nocompress]{cite}
\else
  \usepackage{cite}
\fi

\usepackage[pdftex]{graphicx}
\usepackage{eso-pic}
\usepackage{xspace}
\usepackage{array}
\usepackage{times}
\usepackage{epsfig}
\usepackage{amssymb}
\usepackage{pifont}
\usepackage{graphicx}
\usepackage{amsmath}
\usepackage{amssymb}
\usepackage{tabularx}
\usepackage{algorithm}
\usepackage{algorithmicx}
\usepackage{algpseudocode}
\usepackage{amsmath}
\usepackage{url}
\usepackage{multirow}
\usepackage{amsthm}
\usepackage{mathrsfs}
\usepackage{bm}
\usepackage{lipsum}
\usepackage{subfigure}
\usepackage[section]{placeins}
\usepackage{booktabs}
\usepackage{colortbl}

\usepackage{footmisc}
\input{def_commond}
\usepackage[backref]{hyperref} 
\hypersetup{
colorlinks=true,
linkcolor=black,
citecolor=black,
urlcolor=black,
}
\begin{document}
\title{MVSS-Net: Multi-View Multi-Scale Supervised Networks for Image Manipulation Detection}
\input{author}

\markboth{IEEE TRANSACTIONS ON PATTERN ANALYSIS AND MACHINE INTELLIGENCE,~Vol.~x, No.~x, June~2022}%
{Shell \MakeLowercase{\textit{et al.}}: Bare Demo of IEEEtran.cls for Computer Society Journals}

\IEEEtitleabstractindextext{%
\begin{abstract}

\lxr{As manipulating images by copy-move, splicing and/or inpainting may lead to misinterpretation of the visual content, detecting these sorts of manipulations is crucial for media forensics. Given the variety of possible attacks on the content, devising a generic method is nontrivial. Current deep learning based methods are promising when training and test data are well aligned, but perform poorly on independent tests. Moreover, due to the absence of authentic test images, their image-level detection specificity is in doubt. The key question is how to design and train a deep neural network capable of learning generalizable features \emph{sensitive} to manipulations in novel data, whilst \emph{specific} to prevent false alarms on the authentic. We propose multi-view feature learning to jointly exploit tampering boundary artifacts and the noise view of the input image. As both clues are meant to be semantic-agnostic, the learned features are thus generalizable. For effectively learning from authentic images, we train with multi-scale (pixel / edge / image) supervision. We term the new network MVSS-Net and its enhanced version MVSS-Net++. Experiments are conducted in both within-dataset and cross-dataset scenarios, showing that MVSS-Net++ performs the best, and exhibits better robustness against JPEG compression, Gaussian blur and screenshot based image re-capturing.}
\end{abstract}
}

\maketitle
\IEEEdisplaynontitleabstractindextext
\IEEEpeerreviewmaketitle

\section{Introduction}\label{sec:introduction}
\input{introduction}

\section{Related Work}\label{sec:relate}
\input{related}

\section{Proposed Model}\label{sec:method}
\input{method}


\section{Experiments}\label{sec:eval}
\subsection{Experimental Setup}

\input{exp-setup}

\subsection{Ablation Study}\label{ssec:ablation}
\input{exp-ablation}

\subsection{Comparison with State-of-the-art} \label{ssec:eval-sota}

\input{exp-sota}


\section{Conclusions}
\input{conclusion}


\appendix
\input{appendix}

\input{acknowledgment}

\bibliographystyle{IEEEtran}
\bibliography{egbib}

\input{biography}

\end{document}

%% file: def_commond.tex
\usepackage{xspace}
\makeatletter
\DeclareRobustCommand\onedot{\futurelet\@let@token\@onedot}
\def\@onedot{\ifx\@let@token.\else.\null\fi\xspace}

\def\eg{\emph{e.g}\onedot} 
\def\ie{\emph{i.e}\onedot} 
 
\def\etc{\emph{etc}\onedot} 
\def\wrt{w.r.t\onedot} 
\def\etal{\emph{et al}\onedot}
\makeatother

\newcommand{\del}[1]{\textcolor{red}{\sout{}}}

\newcommand{\dcb}[1]{\textcolor{black}{#1}}
\newcommand{\lxr}[1]{\textcolor{black}{#1}}
\definecolor{mycyan}{cmyk}{.10,.075,.07,0}

\newcommand{\reducedstrut}{\vrule width 0pt height 1.1\ht\strutbox depth 1.1\dp\strutbox\relax}
\newcommand{\incolor}[1]{%
  \begingroup
  \setlength{\fboxsep}{0pt}%
  \colorbox{mycyan}{\reducedstrut\textbf{#1}\/}%
  \endgroup
}

\def \model {\textit{MVSS-Net}}
\def \modelplus {\textit{MVSS-Net++}}

\definecolor{darkgreen}{RGB}{50,150,50}

\newcommand{\specialcell}[2][c]{%
  \begin{tabular}[#1]{@{}c@{}}#2\end{tabular}}

\DeclareMathOperator*{\argmax}{arg\,max}

%% file: author.tex
\author{Chengbo~Dong*,~Xinru~Chen*,~Ruohan~Hu,~Juan~Cao and Xirong~Li,~\IEEEmembership{Member,~IEEE,}

\IEEEcompsocitemizethanks{\IEEEcompsocthanksitem 
Chengbo~Dong, Xinru~Chen, Ruohan~Hu and Xirong~Li are with the Key Lab of Data Engineering and Kowledge Engineering, Renmin University of China and the AIMC Lab, School of Information, Renmin University of China, Beijing 100872, China.\protect\\
E-mail: dongchengbo@ruc.edu.cn, chen\_xinru1999@163.com,  huruohan1126@163.com, xirong@ruc.edu.cn
\IEEEcompsocthanksitem Juan Cao is with Institute of Computing Technology, Chinese Academy of Sciences and the the Key Laboratory of Media Convergence Production Technology and Systems, Beijing 100864, China.\protect\\
E-mail: caojuan@ict.ac.cn
\IEEEcompsocthanksitem{Chengbo Dong and Xinru Chen contributed equally to this work. Corresponding author: Xirong Li.}\protect\\}
}

%% file: introduction.tex





\IEEEPARstart{D}{igital} images can now be manipulated with ease and often in a visually imperceptible manner \cite{Gafni_2020_CVPR}. \emph{Copy-move} (copy and move elements from one region to another region in a given image), \emph{splicing} (copy elements from one image and paste them on another image) and \emph{inpainting} (removal of unwanted elements) are three common types of image manipulation that could lead to misinterpretation \lxr{and thus malicious use} of the visual content \cite{JLSTM,H-LSTM,Mahfoudi2019DEFACTO,review2}. \lxr{Auto-detection of the presence of these sorts of manipulations in a given image is crucial for media forensics and trustworthy information sharing in the cyberspace}. 
We aim to not only discriminate manipulated images from the authentic, but also pinpoint tampered regions at the pixel level.

\input{fig-opening}

\input{fig-model}

\lxr{While pictorial content tampering has been long existing, media forensics is a relatively new research field \cite{review2}. Traditionally, carefully hand-crafted features are extracted from a given image to capture subtle differences between its tampered and authentic regions. The differences are calculated by varied approaches, including media-format based compression artifacts \cite{tradition_jpeg1,tradition_jpeg2}, physics-based lighting inconsistency \cite{tradition_illumination1,tradition_illumination2}, statistical modeling \cite{tradition_statistic}, local noise estimation \cite{tradition_noise}, \etc. However, due to the variety of possible attacks on the digital content, a major challenge in the field is that manipulation detection may not be resolved by a single approach with a single source of information. What makes the problem even more challenging is that when images are uploaded and circulate on social media platforms, regular low-level image processing such as re-sizing, re-compression, re-capturing and aesthetic image enhancement, inevitably weakens forensic traces \cite{Pasquini_2021}.
Towards conquering the challenges, unsurprisingly}, the state-of-the-arts  are deep learning based \cite{2017MFCN,2018rgbn,mantranet,2020GSR,2020Constrained,2020SPAN}, specifically focusing on pixel-level manipulation detection \cite{2017MFCN,mantranet,2020GSR}, \lxr{also known as manipulation localization \cite{Cozzolino2022}}. With only two classes (\emph{manipulated} versus \emph{authentic}) in consideration, the task appears to be a simplified case of image semantic segmentation.
However, an off-the-shelf semantic segmentation network is suboptimal for the task, as it is designed to capture semantic information, making the network dataset-dependent and do not generalize. Prior research \cite{2020GSR} reports that DeepLabv2 \cite{deeplabv2} trained on the CASIAv2 dataset \cite{casiav2} performs well on the CAISAv1 dataset \cite{casiav1} homologous to CASIAv2, yet performs poorly on the non-homologous COVER dataset \cite{2016COVERAGE}. A similar behavior of FCN \cite{fcn} is also observed in this study. \lxr{It has been increasingly recognized that deep neural networks (DNNs) perform well when the training and test data are well aligned in terms of their data source and manipulation methods, but often perform badly on independent tests \cite{Cozzolino2022}.}
Hence, the key question is how to design and train a DNN capable of learning \emph{semantic-agnostic} features  \emph{sensitive} to manipulations, whilst \emph{specific} to prevent false alarms?

\input{table/table-taxonomy}

In order to learn semantic-agnostic features, image content \lxr{originally presented in the RGB view} has to be suppressed. Depending on at what stage the suppression occurs, we categorize existing methods into two groups, \ie noise-view methods \cite{2018rgbn,mantranet,HPFCN,2020Constrained,2020SPAN} and edge-supervised methods \cite{2017MFCN,2020GSR}. Given the hypothesis that novel elements introduced by splicing and/or inpainting differ from the authentic part in terms of their noise distributions, the \lxr{noise-view} methods aim to exploit such discrepancy. A noise map of an input image, generated either by pre-defined high-pass filters\cite{2012SRM} or by their trainable counterparts~\cite{Bayar-journal,HPFCN}, is fed into a \lxr{DNN}, either alone \cite{HPFCN,2020Constrained} or together with the input image \cite{2018rgbn,mantranet,2020SPAN}. Note that the methods are ineffective for detecting copy-move which introduces no new element. The \lxr{edge-supervised} methods try to find boundary artifacts around a tampered region,
implemented by using \lxr{an object-detection head to regress a bounding box to cover the region \cite{2018rgbn, 2020Constrained} or} an auxiliary branch to reconstruct the region's edge \cite{2017MFCN,2020GSR}. 
%
Note that the prior arts uniformly sum \cite{2017MFCN} or concatenate \cite{2020GSR} features from different layers of the backbone as input of the auxiliary branch. As such, there is a risk that deeper-layer features, which are responsible for manipulation detection, remain semantic-aware and thus not generalizable. 

To measure a model's generalizability, a common evaluation protocol \cite{2017MFCN,2020GSR,mantranet,2020SPAN} is to first train the model on a public dataset, say CASIAv2 \cite{casiav2},  and then test it on other public datasets such as NIST16 \cite{NIST}, Columbia \cite{HsuColumbia}, and CASIAv1 \cite{casiav1}. To our surprise, however, the evaluation is performed exclusively on manipulated images, with pixel-level metrics reported. The specificity of the model, which reveals how it handles authentic images and is thus crucial for real-world usability, is ignored. As shown in Fig. \ref{fig:intro-samples}, \lxr{both traditional Error Level Analysis (ELA) and current deep learning methods \cite{2017MFCN,mantranet,2020SPAN} make} serious false alarms on authentic images. \lxr{A}s the current methods mainly use pixel-wise segmentation losses to which an authentic example can contribute is  marginal, it is difficult for these methods to \lxr{exploit the authentic data as holistic context} to improve their specificity.

\lxr{Given the need of exploiting the noise view along with the original RGB view and the need of jointly considering both local edge information and holistic context, it is nontrivial to design a DNN that performs the manipulation detection task in general.} 
We get inspiration from contemporary advances in other \lxr{research domains}. In the context of generic semantic segmentation, the Border Network \cite{border} aggregates features progressively to predict object boundaries. We adapt that technique for tracing subtle boundary artifacts around manipulated regions. \lxr{In the context of medical image analysis}, LesionNet \cite{lesion} incorporates an image classification loss for segmenting retinal lesions in color fundus photographs. We borrow this idea to take authentic images into account. 
We propose \textit{multi-view} feature learning with \emph{multi-scale} supervised networks (MVSS-Net series) for image manipulation detection. 
\lxr{Note that several previous approaches can potentially and indirectly learn the boundary artifacts along with the noise view, \eg via a bounding-box regression task \cite{2018rgbn}.}
To the best of our knowledge (Table \ref{tabel:related}), we are the first to jointly exploit the noise view and the \lxr{\textit{explicitly} extracted} boundary artifacts to learn manipulation detection features. \lxr{With multi-scale supervision}, we also make an initial endeavor to learn from the authentic data. Note that the above joint exploitation is technically nontrivial. For instance, simply adding the image classification loss improves the model specificity, but at the cost of considerable degrade in pixel-level detection performance, as our experiments show. To combine the best of the two worlds, new networks are needed.

\lxr{To sum up}, our \lxr{major} contributions are as follows: \\
$\bullet$ \textbf{\lxr{Proposed} \model~as a new network for image manipulation detection}.  As  Fig. \ref{fig:model} shows, \lxr{the technical strength of \model~lies in its capability to jointly exploit the multi-view input, the explicitly extracted boundary artifacts and the holistic information in an end-to-end manner. Multi-view feature learning is designed to extract} semantic-agnostic and thus more generalizable features. \\
$\bullet$ 
\textbf{\lxr{Network training by} multi-scale supervision}. \lxr{This} allows us to learn \lxr{effectively} from authentic images, which are ignored by the prior arts. Consequently, the \lxr{manipulation detection}  specificity \lxr{is improved} substantially. \\
$\bullet$ \textbf{\lxr{Superior to the SOTA on multiple benchmarks}}.  As extensive experiments on two training sets and six test sets show, \model~compares favorably against the SOTA. \lxr{The inclusion of authentic test images reveals a model's detection specificity at the image level.} Code and models are available at  GitHub\footnote{\label{github}\url{https://github.com/dong03/MVSS-Net}}.

A preliminary version of this work was published at ICCV 2021 \cite{iccv21-mvssnet}. The journal article improves over the conference paper in multiple aspects. First, for converting pixel-level manipulation detection to an image-level prediction, we propose ConvGeM to replace global max pooling (GMP) used in \cite{iccv21-mvssnet}. The new module effectively overcomes two downsides of GMP, \ie the bottleneck in back propagating the image-scale loss and the lack of ability to consider the amount and the spatial distribution of positive responses. This results in a better model \modelplus. Second, we strengthen our evaluation by including three more baseline methods, \ie H-LSTM \cite{H-LSTM}, SPAN \cite{2020SPAN} and CAT-Net \cite{2021CAT}, and a recently released dataset, \ie IMD \cite{2020IMD2020}. In addition, we present a pilot study on how the current models react to manipulated images given re-capturing by screenshot, a common operation when images circulate on the Internet.

%% file: fig-opening.tex
\begin{figure} 
    \begin{center}
    \includegraphics[width=\columnwidth]{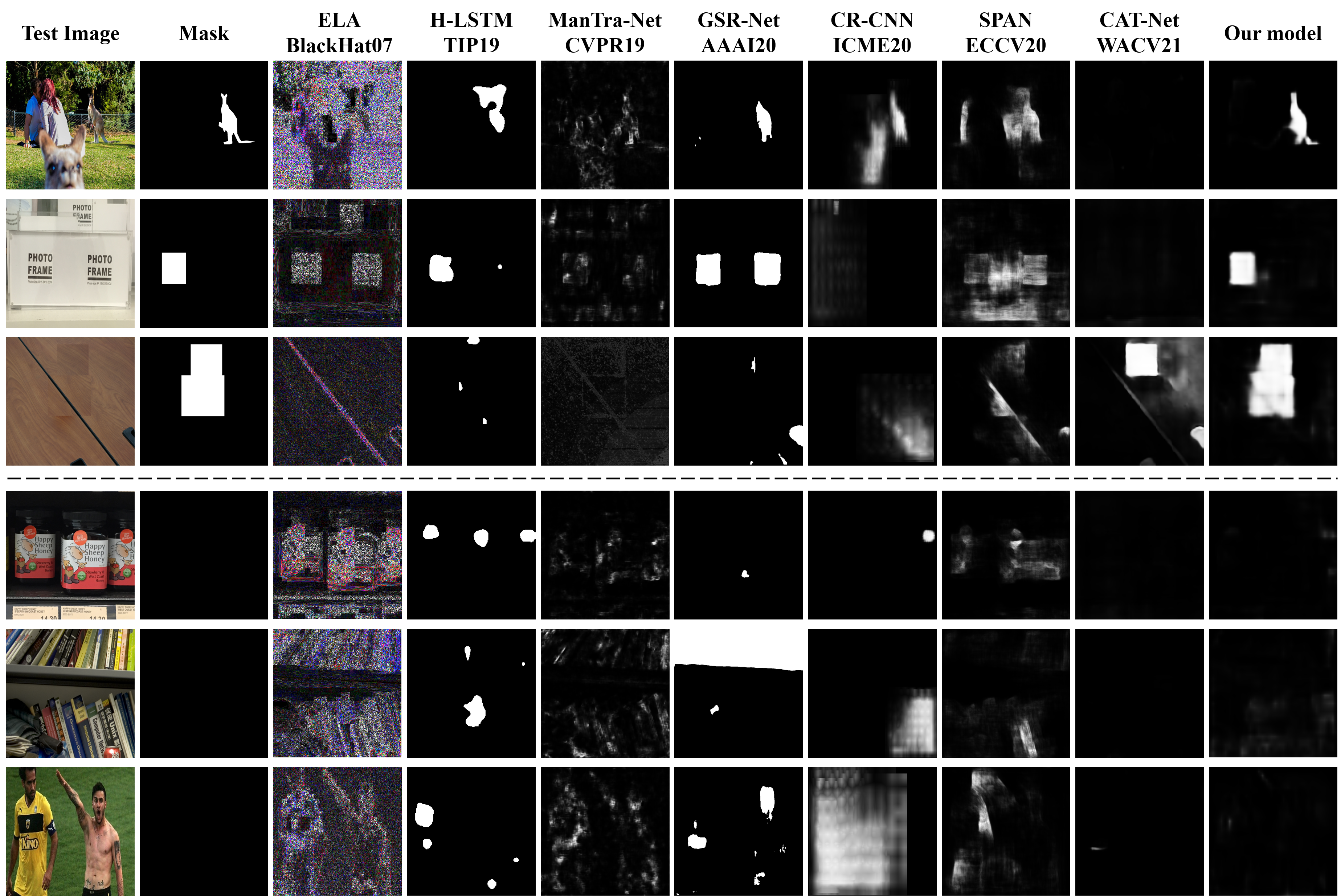}
    \end{center}
    \caption{\textbf{Image manipulation detection by the state-of-the-art}. Test images in the first three rows are manipulated by splicing, copy-move and inpainting, respectively. Test images in the last three rows are authentic (thus with blank mask). Our model (\modelplus) strikes a good balance between detection sensitivity (lower miss detection on the manipulated) and specificity (lower false alarm on the authentic).}
    \label{fig:intro-samples}
\end{figure}

%% file: fig-model.tex
\begin{figure*}[htpb]
    \begin{center}
    \includegraphics[width=\textwidth]{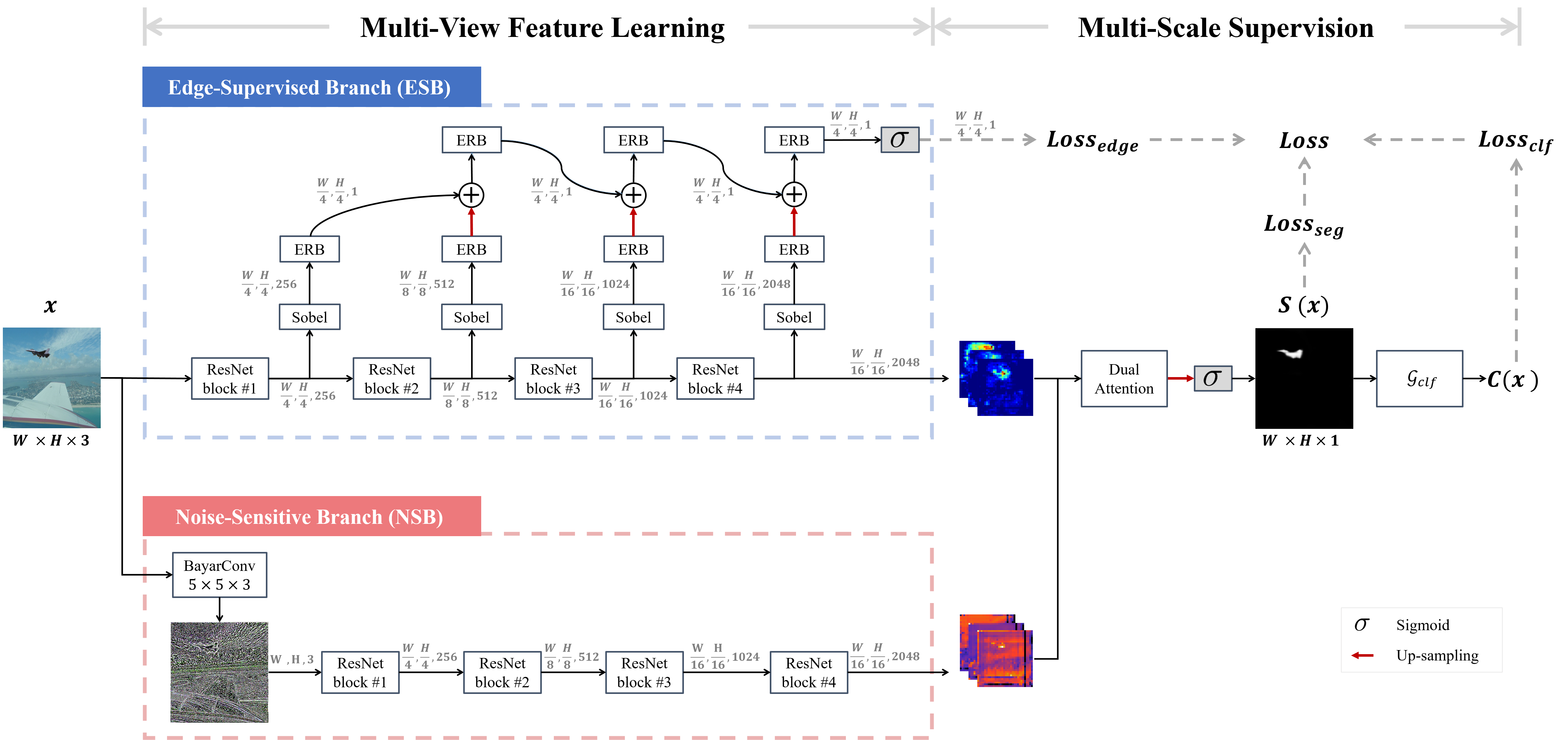}
    \end{center}
    \caption{\textbf{Conceptual diagram of the proposed multi-view multi-scale supervised networks for image manipulation detection}. We use the edge-supervised branch (ESB) and the noise-sensitive branch (NSB) to learn semantic-agnostic features for manipulation detection, and multi-scale supervision to strike a proper balance between model sensitivity and specificity. The non-trainable sigmoid ($\sigma$) layer is shown in gray. The $\mathcal{G}_{clf}$ module is responsible for converting a pixel-level segmentation map $S(x)$ to an image-level prediction $C(x)$.  Depending on how  the module is implemented, we have \model ~which uses global max pooling (GMP) and \modelplus ~which uses ConvGeM.}
    \label{fig:model}
\end{figure*}

%% file: table/table-taxonomy.tex
\begin{table*}[htpb!]
\caption{\textbf{A taxonomy of the state-of-the-art for image manipulation detection}. Methods marked with $\dagger$ are  open-sourced or with  models released, and will be compared in our experiments. \lxr{Star (*) indicates edge information is implicitly considered via bounding-box regression}. Edge and image labels used by this paper are automatically extracted from pixel-level annotations. So our multi-scale supervision requires no extra manual labeling.}
\renewcommand{\arraystretch}{1}
\setlength\tabcolsep{3.5pt}
\begin{center}
\scalebox{0.95}{

\begin{tabular}{lccclp{0.7cm}<{\centering}p{0.7cm}<{\centering}p{0.7cm}<{\centering}p{0.05cm}p{1cm}<{\centering}p{1cm}<{\centering}}
\toprule
\multicolumn{1}{c}{\multirow{2}{*}{\textbf{Method}}} &
  \multicolumn{3}{c}{\textbf{Views}} &
  \multicolumn{1}{l}{\multirow{2}{*}{\textbf{\specialcell{Semantic\\ segmentation\\ backbone}}}} &
  \multicolumn{3}{c}{\textbf{Scales of supervision}} & & \multicolumn{2}{c}{\textbf{Level of evaluation}}\\ \cmidrule{2-4} \cmidrule{6-8} \cmidrule{10-11} 
\multicolumn{1}{c}{} &
  \textit{RGB} &
  \textit{Noise} &
  \textit{Fusion} &
  \multicolumn{1}{c}{} &
  \textit{pixel} &
  \textit{edge} &
  \textit{image} & &
  \textit{pixel} &
  \textit{image}\\ \midrule

MFCN, Salloum \etal 2017\cite{2017MFCN}        & + & -                & -                                                                   & FCN        & + & + & - &\specialcell{~\\~}& + & - \\ \hline

RGB-N, Zhou \etal 2018\cite{2018rgbn}       & + & SRM filter       & \begin{tabular}[c]{@{}c@{}}late fusion\\ (bilinear pooling)\end{tabular} & Faster R-CNN & - & * & -& & + &-\\ \hline

H-LSTM$\dagger$, Bappy \etal 2019\cite{H-LSTM}        & + & -            & -                                                                        & Patch-LSTM & + & - & - &\specialcell{~\\~}&+ & - \\ \hline
ManTra-Net$\dagger$, Wu \etal 2019\cite{mantranet} &
  + &\specialcell{SRM filter\\ BayarConv} &
  \specialcell{early fusion\\ (feature concatenation)} &
  Wider VGG &
  + &
  - & 
  - & &+ & -\\ \hline
HP-FCN$\dagger$, Li \& Huang 2019\cite{HPFCN}     & - & High-pass filters & -                                                                    & FCN        & + & - & - &\specialcell{~\\~}& + & -\\ \hline

GSR-Net$\dagger$, Zhou \etal 2020\cite{2020GSR}      & + & -                & -                                                                        & DeepLabv2  & + & + & -&\specialcell{~\\~} & + & -\\ \hline
CR-CNN$\dagger$, Yang \etal 2020\cite{2020Constrained}        & - & BayarConv  & -                                                                      & Mask R-CNN & + & * & - &\specialcell{~\\~}& + & - \\ \hline
SPAN$\dagger$, Hu \etal 2020\cite{2020SPAN}         & + &\begin{tabular}[c]{@{}c@{}}SRM filter\\ BayarConv\end{tabular} 
&  \specialcell{early fusion\\ (feature concatenation)}    & Wider VGG  & + & - & - &&+ & -\\ \hline

MM-Net, Yang \etal 2021\cite{MM-Net}        & + & BayarConv  & \specialcell{middle fusion\\ (attention guidance)}                                                                        & Mask R-CNN & + & - & - && + & - \\ \hline


JPEG-ComNet, Rao \& Ni 2021\cite{RaoJPEG}        & + & SRM filter  & \specialcell{early fusion\\ (feature concatenation)}                                                                        & Siamese FCN & + & + & - && + & -\\ \hline

CAT-Net$\dagger$, Kwon \etal 2021\cite{2021CAT}        & + & DCT  & \specialcell{middle fusion\\ (feature concatenation)}                                                                        & HRNet & + & - & - & &+ & -\\ \hline

Proposed \model$\dagger$ & + & BayarConv      & \specialcell{late fusion\\ (dual attention)}   & FCN        & + & + & + && + & + \\ \bottomrule
\end{tabular}%
}
\end{center}

\label{tabel:related}
\end{table*}

%% file: related.tex


\lxr{We are} inspired by a number of recent works that made novel attempts to learn semantic-agnostic features for image manipulation detection, see Table \ref{tabel:related}. 
We describe in brief how these attempts are implemented and explain our novelties accordingly. We focus on deep learning approaches to copy-move / splicing / inpainting detection. For the detection of low-level manipulations such as Gaussian Blur and JPEG compression, we refer to \cite{Bayar-journal}.

In order to suppress the content information, Li and Huang \cite{HPFCN} propose to implement an FCN's first convolutional layer with trainable high-pass filters and apply their HP-FCN for inpainting detection. Kown \etal \cite{2021CAT} model quantized DCT coefficient distribution to trace compression artifacts. Yang \etal use BayarConv\cite{Bayar-journal} as the initial convolutional layer of their CR-CNN \cite{2020Constrained}. 
Although such constrained conv. layers are helpful for extracting noise information, using them alone has the risk of losing other useful information in the original RGB view. Hence, we see an increasing number of works on exploiting information from both the RGB view and the noise view \cite{2018rgbn,mantranet,2020SPAN,RaoJPEG,MM-Net,2021CAT}. 
Zhou \etal \cite{2018rgbn} develop a two-stream Faster R-CNN, coined RGB-N, which takes as input the RGB image and its noise counterpart generated by the spatial rich model (SRM)\cite{2012SRM}. 
Rao and Ni also use SRM \cite{RaoJPEG}, whilst Wu \etal \cite{mantranet} and Hu \etal \cite{2020SPAN} use both BayarConv and SRM. 
Given features from distinct views, the need for feature fusion is on. 
Feature concatenation at an early stage is adopted by Mantra-Net \cite{mantranet}, SPAN \cite{2020SPAN} and JPEG-ComNet \cite{RaoJPEG}, while CAT-Net \cite{2021CAT} concatenates the features at a middle stage. Alternatively, MM-Net \cite{MM-Net} performs feature fusion at an intermediate stage, where features from the noise-view branch are used as attention maps to re-weight features from the RGB-view branch. 
Our \model~is more close to RGB-N as it performs feature fusion at the late stage. However, different from the non-trainable bilinear pooling used in RGB-N, Dual Attention used in \model~is trainable and thus more selective.



As manipulating a specific region in a given image inevitably leaves traces between the tampered region and its surrounding, how to exploit such edge artifact also matters for manipulation detection. Salloum \etal develop a multi-task FCN (MFCN) to symmetrically predict a tampered area and its boundary \cite{2017MFCN}. 
GSR-Net has an edge detection and refinement branch which accepts features from different levels\cite{2020GSR}. The more recent JPEG-ComNet \cite{RaoJPEG} applies boundary attention on RGB view features to predict edges of manipulated areas, and subsequently utilizes the prediction to refine manipulation segmentation. Given that region segmentation and edge detection are intrinsically two distinct tasks, the challenge lies in how to strike a proper balance between the two. Directly using deeper features for edge detection as done in JPEG-ComNet has the risk of affecting the main task of manipulation segmentation, while putting all features together as used in MFCN and GSR-Net may let the deeper features be ignored by the edge branch. Our \model~has an edge-supervised branch that effectively resolves these issues.

Last but not least, we observe that the specificity of an image manipulation detector, \ie how it responds to authentic images, is seldom reported. In fact, the mainstream solutions are developed within a  semantic segmentation network. Naturally, they are trained and evaluated on manipulated images in the context of manipulation segmentation \cite{2020GSR}. The absence of authentic images both in the training and test stages naturally raises concerns regarding the detection specificity. In this paper we make a novel attempt to include authentic images for training and test, an important step towards real-world deployment. In addition, different from the previous common practice that selects a model's decision threshold based on test data, we advocate the use of a default threshold of $0.5$. 
Such an evaluation also matters practically.




%% file: method.tex

Given an RGB image $x$ of size $W\times H \times 3$, we aim for a multi-head deep network $\mathcal{G}$ 
that not only determines whether the image has been manipulated, but also pinpoints its manipulated pixels. In particular, we let  $\mathcal{G}$ have an semantic segmentation head, denoted by $\mathcal{G}_{seg}$, for  producing  a full-size probability map, denoted by $S(x)$, which indicates the probability of manipulation at the pixel level. We have access to the pixel-level scores via $S_{i,j}(x)$, $i=1,\ldots,W, j=1,\ldots,H$. Meanwhile, the network has an image classification head $\mathcal{G}_{clf}$ to output $C(x)$ the probability of the image being manipulated.
As the image-level decision is naturally subject to pixel-level evidence, we  derive $C(x)$ from the segmentation map:
\begin{equation} \label{eq:general}
\left\{ 
\begin{array}{ll}
S(x) & \leftarrow \mathcal{G}_{seg}(x),\\
C(x) & \leftarrow \mathcal{G}_{clf}(S(x)).
\end{array} \right.
\end{equation}
Eq. \ref{eq:general} provides a high-level sketch of our network. 

In order to extract generalizable manipulation detection features, $\mathcal{G}$ is designed to accept both the original RGB-view and an extra noise-view of the input image. To strike a proper balance between detection sensitivity and specificity, the multi-view feature learning process is jointly supervised by annotations of three scales, \ie pixel, edge and image. 
All this results in Multi-View multi-Scale Supervised Networks (\model).

\input{method-multiview}

\input{method-ConvGeM}

\input{method-multiscale}

%% file: method-multiview.tex
\subsection{Multi-View Feature Learning} \label{ssec:mvfl}

\model~has two branches, both with ResNet-50 \cite{resnet} as their backbones.
The edge-supervised branch (ESB) at the top of Fig. \ref{fig:model} is specifically designed to exploit subtle boundary artifacts around tampered regions, whilst the noise-sensitive branch (NSB) at the bottom is to capture the noise inconsistency between tampered and authentic regions. Both clues are meant to be semantic-agnostic. 

\subsubsection{Edge-Supervised Branch }\label{sssec:esb}

Ideally, with edge supervision, we hope the response area of the network will be more concentrated on tampered regions. Designing such an edge-supervised network is nontrivial. As noted in Sec. \ref{sec:relate}, the main challenge is how to construct an appropriate input for the edge detection head. On one hand, directly using features from the last ResNet block is problematic, as this will enforce the deep features to capture low-level edge patterns and consequently affect the main task of manipulation segmentation. While on the other hand, using features from the initial blocks is also questionable, as subtle edge patterns contained in these shallow features can vanish with ease after multiple deep convolutions. A joint use of both shallow and deep features is thus necessary. However, we argue that simple feature concatenation as previously used in \cite{2020GSR} is suboptimal, as the features are mixed and there is no guarantee that the deeper features will receive adequate supervision from the edge head. To conquer the challenge, we propose to construct the input of the edge head in a shallow-to-deep manner.

\begin{figure}[htbp]
\begin{center}
\subfigure[Sobel Layer]{
\begin{minipage}[t]{0.90\linewidth}
\begin{center}
\includegraphics[width=0.9\columnwidth]{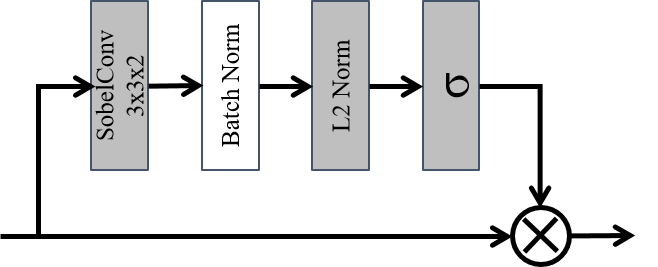}
\end{center}
\label{fig:sobel}
\end{minipage}%
}%

\subfigure[Edge Residual Block (ERB)]{
\begin{minipage}[t]{0.95\linewidth}
\begin{center}
\includegraphics[width=0.95\columnwidth]{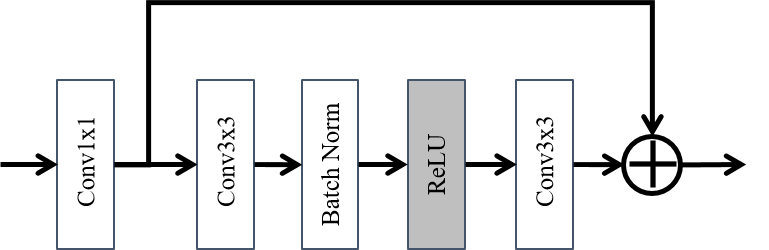}
\end{center}
\label{fig:erb}
\end{minipage}%
}%

\end{center}
\caption{\textbf{Diagrams of (a) Sobel layer and (b) edge residual block}, used in ESB for manipulation edge detection.}
\label{fig:sobel-erb}
\end{figure}

As illustrated in Fig. \ref{fig:model}, features from different ResNet blocks are combined in a progressive manner for manipulation edge detection. 
In order to enhance edge-related patterns, we introduce a Sobel layer, see Fig. \ref{fig:sobel}. 
The basic idea behind the Sobel layer is to discriminate edge-related pixels from others in a given feature map by attending to them with edge-related weights. In order to obtain such an attention map, we let the feature map go through the classical Sobel filter, which is widely used for identifying candidate edge pixels \cite{gonzalez2009digital}, followed by a Batch Normalization layer and an L2 Norm layer, and eventually a sigmoid ($\sigma$) layer. The feature map is then re-weighted using the attention map with element-wise multiplication.

\begin{figure}[htpb]
    \begin{center}
    \includegraphics[width=\columnwidth]{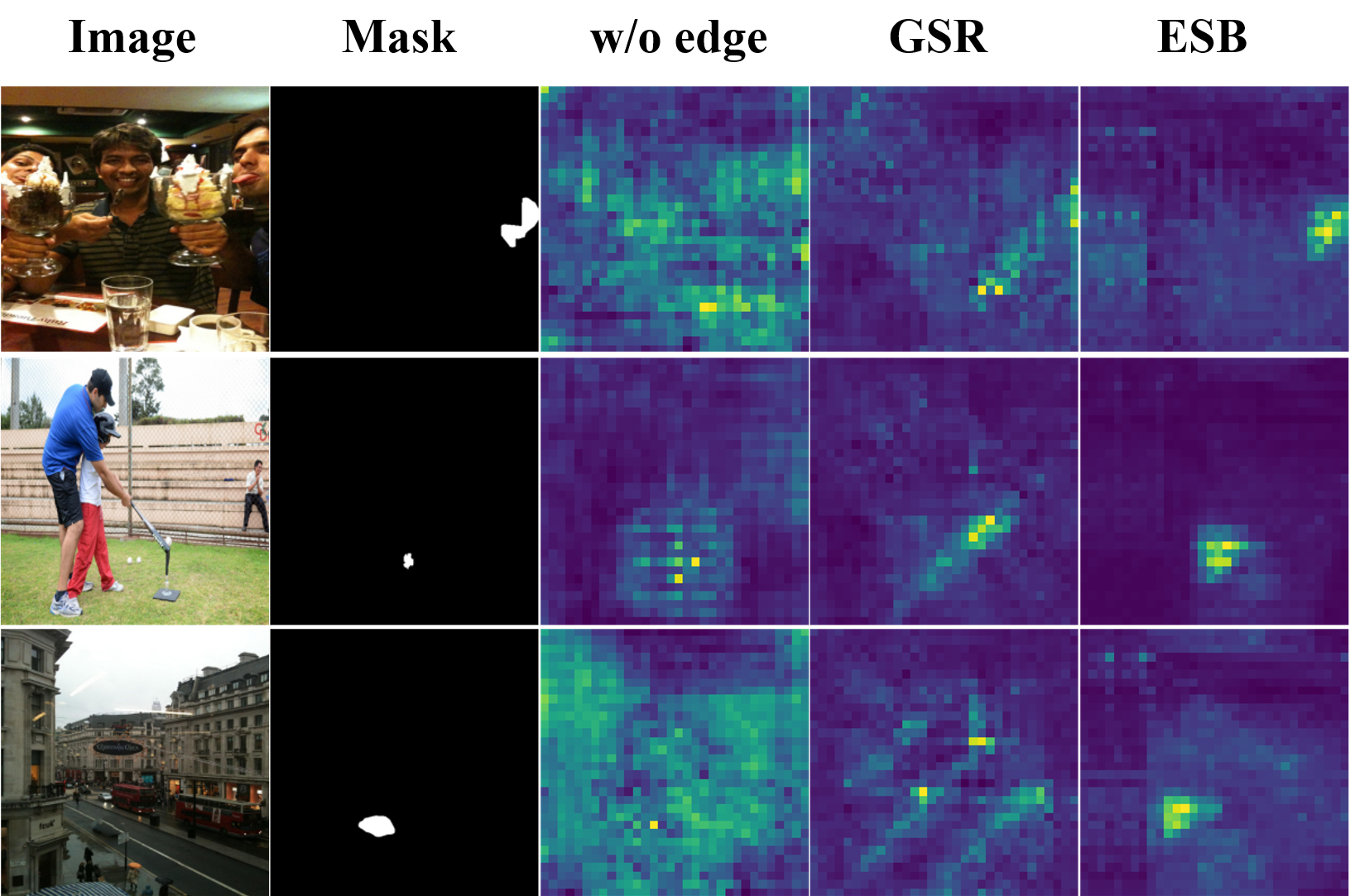}
    \end{center}
    \caption{\textbf{Visualization of averaged feature maps of the last ResNet block}, brighter color indicating higher responses.  Manipulation from the top to bottom is inpainting, copy-move and splicing. Read from the third column  are \textit{w/o edge} (standard ResNet with no edge-related block), \textit{GSR} (ResNet with the GSR-Net alike edge branch) and our \textit{ESB}, which produces more focused responses near tampered regions.}
    \label{fig:visual_esb}
\end{figure}

The feature map produced by the block \#$i$, enhanced by the Sobel layer, then goes through an edge residual block (ERB), see Fig. \ref{fig:erb}, before being combined (by summation) with  its counterpart from the block \#$i$+1. To prevent the effect of accumulation that unwittingly makes features from the last blocks slighted, we let the combined features go through another ERB (top in Fig. \ref{fig:model}) before the next round of feature combination. We believe such a mechanism helps prevent extreme cases wherein deeper features are either over-supervised or fully ignored by the edge head. By visualizing feature maps of the last ResNet block in Fig. \ref{fig:visual_esb}, we observe that the proposed ESB indeed produces  more focused responses near tampered regions.

The output of ESB has two parts: feature maps from the last ResNet block, denoted as $\{f_{esb,1}, \ldots, f_{esb,k}\}$, 
to be used for the main tasks, and the predicted manipulation edge map, denoted as $S_{edge}(x)$, obtained by transforming the output of the last ERB with a sigmoid ($\sigma$) layer. The key data flow of the ESB branch is conceptually expressed by Eq. \ref{eq:esb},
\begin{equation}\label{eq:esb}
\left. \begin{array}{*{20}{r}}
\{f_{esb,1}, \ldots, f_{esb,k} \}\\
S_{edge}(x)
\end{array} \right\} \!\leftarrow \!\mbox{ESB}(x).
\end{equation}


\subsubsection{Noise-Sensitive Branch }\label{sssec:nsb}

In parallel to ESB, 
we build a noise-sensitive branch (NSB). 
NSB is implemented as a standard FCN (another ResNet-50 as its backbone) except for its input, which is a noise view of a given image rather than the original RGB view. Regarding the choice of noise extraction, we adopt BayarConv \cite{Bayar-journal}, which is found to be better than the SRM filter \cite{2020Constrained}. 

According to Bayar and Stamm \cite{Bayar-journal}, BayarConv is developed to enhance the noise inconsistency between manipulated and authentic regions within a given image. To that end, the BayarConv layer is implemented as a set of trainable prediction error filters. The response of each filter is designed to be the error between the center-pixel value of the filter window and the linear combination of the remaining pixel values within the window. More concretely, given a specific convolutional filter parameterized by $\omega$ with  $\omega(0,0)$ as its center element, BayarConv imposes two constraints, \ie $\omega(0,0)=-1$ and $\sum\limits_{i,j \ne 0} \omega(i,j)=1$. The constraints are applied on $\omega$ after each training iteration. 

As Fig. \ref{fig:model} shows, letting the given image $x$ go through a BayarConv layer with kernel size of $5\times 5 \times 3$, we obtain its full-sized noise view as $\mbox{BayarConv}(x)$. The output of the NSB branch is an array of $k$ feature maps from the last ResNet block of its backbone, \ie
\begin{equation} \label{eq:nsb}
\{f_{nsb,1}, \ldots, f_{nsb,k}\} \leftarrow \mbox{ResNet}(\mbox{BayarConv}(x)).
\end{equation}

\subsubsection{Branch Fusion by Dual Attention}\label{sssec:fusion}

Given two arrays of feature maps $\{f_{esb,1},\ldots,f_{esb,k}\}$ and $\{f_{nsb,1}, \ldots, f_{nsb,k}\}$ from ESB and NSB, we propose to fuse them by a trainable Dual Attention (DA) module \cite{danet}. This is new, because previous work \cite{2018rgbn} uses bilinear pooling for feature fusion, which is non-trainable.

\begin{figure}[htpb]
    \begin{center}
    \includegraphics[width=\columnwidth]{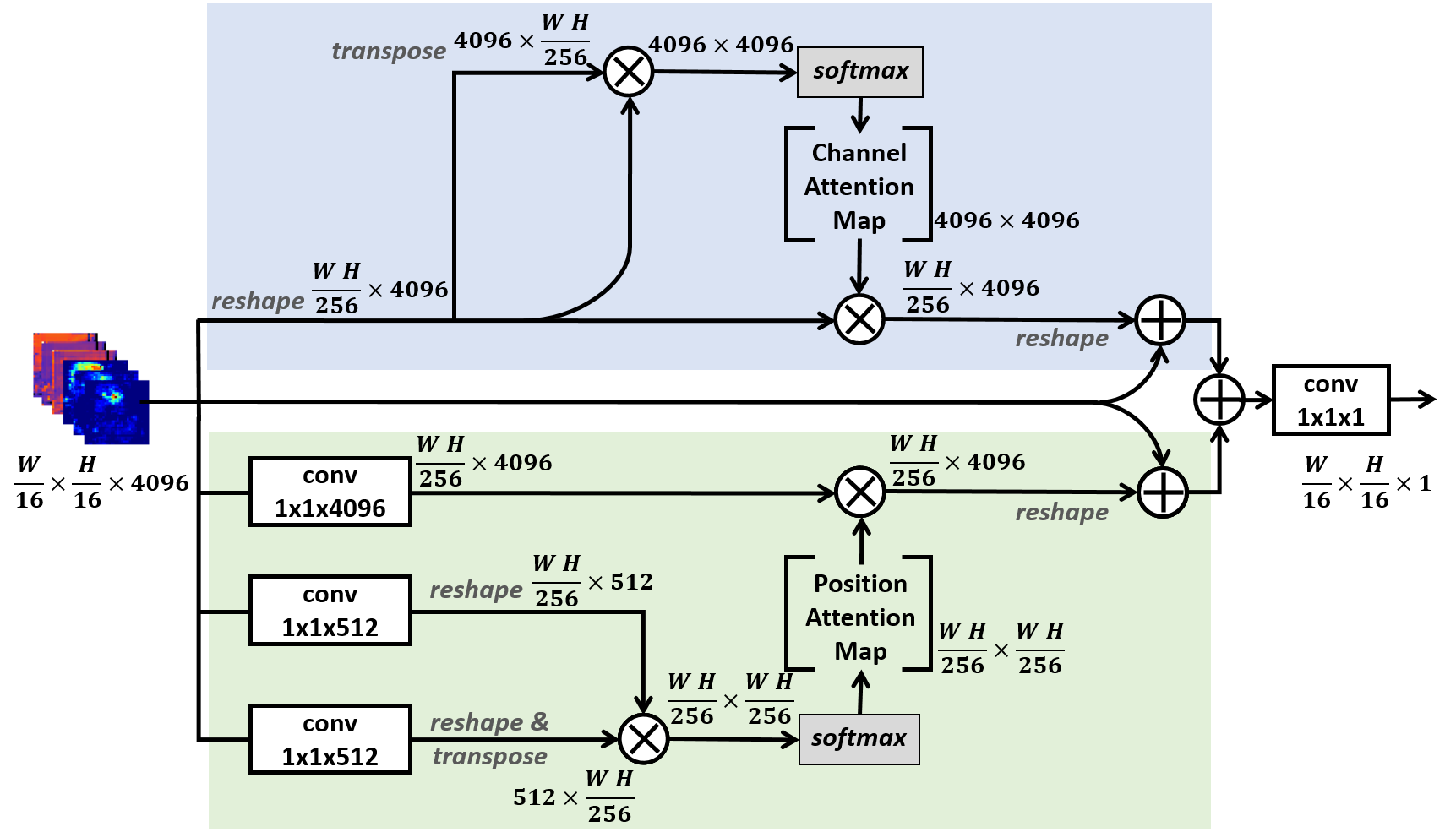}
    \end{center}
    \caption{\textbf{Dual Attention} (DA), with its channel attention module shown in blue and its position attention module shown in green. While DA was originally developed for capturing long-range contextual dependencies of feature maps produced by a single-branch network \cite{danet}, we re-purpose it for fusing feature maps from two distinct branches.}
    \label{fig:da}
\end{figure}

The DA module has two attention mechanisms working in parallel: channel attention (CA) and position attention (PA), see Fig. \ref{fig:da}. CA  associates channel-wise  features to selectively emphasize interdependent channel feature maps. Meanwhile, PA selectively updates features at each position by a weighted sum of the features at all positions. The outputs of CA and PA are summed up, and transformed via a $1 \times 1$ convolution into a feature map of size $\frac{W}{16}\times \frac{H}{16}$, denoted as $S^{'}(x)$. With parameter-free bilinear upsampling followed by an element-wise sigmoid function, $S^{'}(x)$ is transformed into the full-size segmentation map $S(x)$. The DA based branch fusion is conceptually expressed as
\begin{equation}
\left\{\! {\begin{array}{*{5}{l}}
S^{'}\!(x) \!\leftarrow\! \mbox{DA}([f_{esb,1}\!, \!\ldots \!,\!f_{esb,k},\!f_{nsb,1}, \ldots ,f_{nsb,k}]),\\

S(x) \!\leftarrow\! \sigma (\mbox{bilinear-upsampling}( S^{'}(x) )).
\end{array}} \right.\label{eq:da}
\end{equation}

%% file: method-ConvGeM.tex

\subsection{ConvGeM for Image-Level Prediction}\label{ssec:convgem}


Concerning $\mathcal{G}_{clf}$ in Eq. \ref{eq:general}, a straightforward implementation is Global Max Pooling (GMP) as previously used in our conference paper \cite{iccv21-mvssnet}. GMP takes the maximum of $S(x)$ as $C(x)$, \ie $C(x)=S_{i^*,j^*}(x)$, with $(i^*, j^*) = \argmax_{i,j} S_{i,j}(x)$. While GMP links $C(x)$ directly to $S(x)$, we argue that this operation is suboptimal due to the following two downsides. First, as an image classification loss is practically computed based on $S_{i^*,j^*}(x)$, the gradient \wrt the loss is back-propagated exclusively via the sole point $(i^*, j^*)$. Such a bottleneck not only slows down the training of the classification head, but also impedes the head from guiding the entire network.
Second, GMP is invariant to the amount of positive responses and how they are spatially distributed. However, both properties matter for the pixel-level detection result to be meaningful. According to Gestalt theory \cite{koffka1922perception}, humans perceive visual patterns in connection with their spatial context. Following this theory, for an effective deception, a certain amount of pixels in a given image have to be manipulated simultaneously with certain configurations. As such, positive responses occurring sporadically are more likely to be noise than their spatially grouped counterparts. Telling them apart is beyond the capability of GMP.

We notice that Generalized Mean pooling (GeM), originally proposed for image retrieval \cite{Gem}, can be used to overcome the first downside of GMP. As shown in Eq. \ref{eq:gem}, GeM uses a trainable positive parameter $p$ to strike a balance between global mean pooling ($p=1$) and GMP (say a larger $p$ of 100):
\begin{equation} \label{eq:gem}
\mbox{GeM}(S(x)) = \frac{1}{W \times H}(\sum_{i = 1}^W \sum_{j=1}^H  S(x_{i,j})^p)^{\frac{1}{p}}.
\end{equation}
As more pixels contribute to $C(x)$, GeM effectively breaks the bottleneck of GMP in back propagation. We empirically observe that substituting GeM for GMP saves 10 training epochs approximately. 
Nonetheless, GeM remains invariant to the spatial distribution of the positive responses. 

\begin{figure}[htbp]
    \centering
    \includegraphics[width=\linewidth]{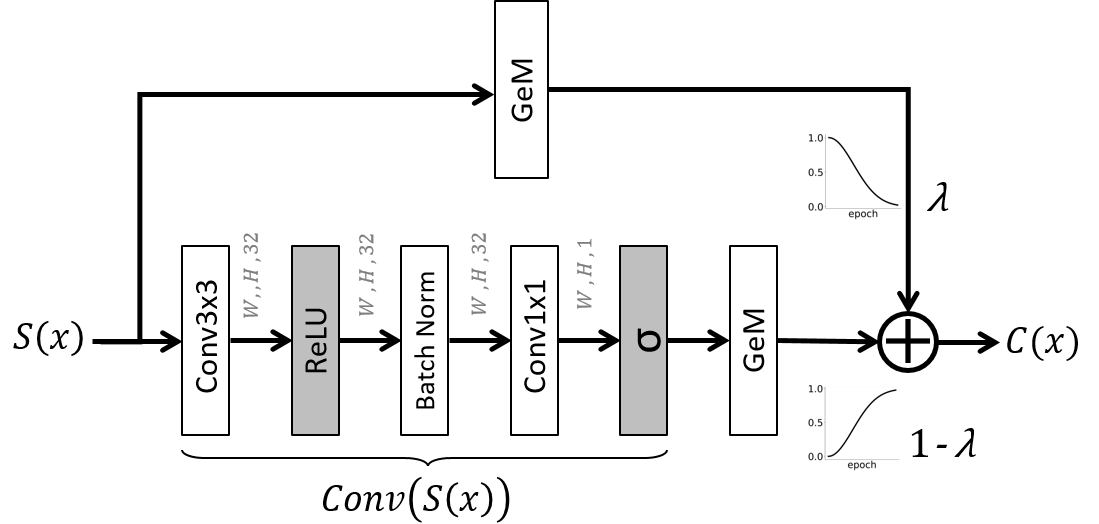}
    
    \caption{\textbf{Illustration of the proposed ConvGeM module} that converts the pixel-level manipulation detection result $S(x)$ to the image-level prediction $C(x)$. The hyper parameter $\lambda$  that balances GeM and GeM(Conv) decays \wrt the training epochs, and is  dynamically determined in the training process.}
    \label{fig:convgem}
\end{figure}

As convolution naturally captures spatial correlation among pixels, one might consider adding a convolutional block, denoted by $\mbox{Conv}(S(x))$, in advance to GeM. Consequently, $C(x)$ is obtained as $\mbox{GeM}(\mbox{Conv}(S(x)))$. Notice that in the early training epochs, the network, in particular its segmentation head $\mathcal{G}_{seg}$, has not been well trained, and thus mostly produces  meaningless $S(x)$. Such noisy input to $\mathcal{G}_{clf}$ will be further exaggerated by $\mbox{Conv}$, making the classification head and consequently the entire network difficult to train. In order to suppress such a negative effect, we add $\mbox{GeM}(S(x))$ to $C(x)$ through a \emph{decayed} skip connection weighed by a nonnegative hyper parameter $\lambda$ as
\begin{equation} \label{eq:convgem}
C(x) = \lambda  \cdot \mbox{GeM}(S(x)) + (1- \lambda) \cdot \mbox{GeM}(\mbox{Conv}(S(x)))
\end{equation}
where $\lambda$ is initialized with a value close to 1, and decayed nonlinearly \wrt the number of epochs.  As illustrated in Fig \ref{fig:convgem},  using a close-to-one $\lambda$ lets $\mathcal{G}_{clf}$  temporarily ignore the Conv block at the early training stage. Then, as $\mathcal{G}_{seg}$ continuously improves to provide more accurate and reliable $S(x)$, $\lambda$ decreases more rapidly to let $\mathcal{G}_{clf}$ count more on $\mbox{Conv}$ to exploit $S(x)$ sufficiently.  As Eq. \ref{eq:convgem} shows, the convex combination of  $\mbox{GeM}$ and $\mbox{GeM}(\mbox{Conv})$ with their weights dynamically determined in the training process effectively tackles the drawbacks of GMP. We coin the new module ConvGeM.

%% file: method-multiscale.tex
\subsection{Multi-Scale Supervision} \label{ssec:loss} 

We consider losses at three scales, each with its own target, \ie a pixel-scale loss for improving the model's sensitivity for pixel-level manipulation detection, an edge loss for learning semantic-agnostic features and an image-scale loss for improving the model's specificity for image-level manipulation detection. 

\textbf{Pixel-scale loss}. As manipulated pixels are typically in minority in a given image, we use the Dice loss, found to be effective for learning from extremely imbalanced data \cite{lesion}:
\begin{equation} 
loss_{seg}(x) = 1 - \frac{2  \sum_{i,j} S(x_{i,j}) \cdot y_{i,i}} {\sum_{i,j} S^2(x_{i,j})  + \sum_{i,j} y^2_{i,j}},
\label{eq:dice}
\end{equation}
where $y_{i,j}\in\{0,1\}$ is a binary label indicating whether pixel $(i,j)$ is manipulated. 

\textbf{Edge loss}. As pixels of an edge are overwhelmed by non-edge pixels, we again use the Dice loss for manipulation edge detection, denoted as $loss_{edg}$. Since manipulation edge detection is an auxiliary task, we do not compute the $loss_{edg}$ at the full size of $W \times H$. Instead, the loss is computed at a smaller size of $\frac{W}{4} \times \frac{H}{4}$, see Fig. \ref{fig:model}. This tactic reduces computational cost during training, and in the meanwhile, improves the performance slightly.

\textbf{Image-scale loss}. In order to reduce false alarms, authentic images have to be taken into account in the training stage. This is however nontrivial for the current works, \eg Mantra-Net \cite{mantranet}, HP-FCN \cite{HPFCN} and GSR-Net \cite{2020GSR}, as they all rely on certain semantic segmentation losses. Consider the widely used binary cross-entropy (BCE) loss for instance. An authentic image with a small percent of its pixels misclassified contributes marginally to the BCE loss, making it difficult to effectively reduce false alarms. Also note that the Dice loss cannot handle the authentic image by definition. Therefore, an image-scale loss is needed. 

As the two classes at the image level are more balanced than their counterpart at the pixel level, we adopt the BCE loss, widely used for image classification, for computing the image-scale loss:
\begin{equation} \label{eq:loss-clf}
loss_{clf}(x) = - ( y \cdot \log C(x)  + (1 - y) \cdot \log(1 - C(x)) ),
\end{equation}
with $y=\max(\{y_{i,j}\})$. 
It is worth pointing out that the usefulness of $loss_{clf}$ is not limited to improving model specificity. Through ConvGeM, the image-scale supervision can now be back propagated more effectively than our previously  used GMP  \cite{iccv21-mvssnet} for improving feature learning. 


\textbf{Combined loss}. Given the losses computed at three distinct scales, we obtain a combined loss by a convex combination, \ie
\begin{equation} \label{eq:loss}
    Loss = \alpha \cdot loss_{seg} + \beta \cdot loss_{clf} + (1 - \alpha - \beta) \cdot loss_{edg}
\end{equation} 
where $\alpha, \beta \in(0,1)$ are positive weights. The combined loss is minimized by stochastic gradient descent, where authentic images in a specific mini-batch  are  used  to compute $loss_{clf}$ only.

%% file: exp-setup.tex


\textbf{Datasets}. For a head-to-head comparison with the state-of-the-art, we adopt CASIAv2 \cite{casiav2} for training and widely used COVER \cite{2016COVERAGE}, Columbia \cite{HsuColumbia}, NIST16 \cite{NIST}, CASIAv1+\footnote{Note that the original CASIAv1 has 782 authentic images in common with CASIAv2. We fixed the issue by replacing these common images in CASIAv1 with the same amount of images randomly sampled from Corel \cite{corel}, which is the data source of CASIAv1. We term the fixed version CASIAv1+.} \cite{casiav1} and the more recent IMD\cite{2020IMD2020} for testing. 
Meanwhile, we notice DEFACTO \cite{Mahfoudi2019DEFACTO}, a recent large-scale dataset, containing 149k images sampled from MS-COCO \cite{Lin2014Microsoft} and auto-manipulated by copy-move, splicing and inpainting. Considering the challenging nature of DEFACTO, we choose to perform our ablation study on this new set. As the set has no authentic images, we construct a training set termed DEF-84k, by randomly sampling 64k positive images from DEFACTO and 20k negative images from MS-COCO. In a similar manner, we build a test set termed DEF-12k, by randomly sampling 6k positive images from the remaining part of DEFACTO and 6k negatives from MS-COCO. Note that to avoid any data leakage, for manipulated images used for training (test), their source images are not included in the test (training) set. 
In total, our experiments use two training sets and six test sets, see Table \ref{table:dataset}. Metadata of these sets is available at our project website\footref{github}.

\input{table/table-datasets}



\textbf{Evaluation Criteria}. 
For pixel-level manipulation detection, following previous works \cite{2017MFCN, 2018rgbn, 2020GSR}, we compute pixel-level precision and recall, and report their F1. 
For image-level manipulation detection, in order to measure the miss detection rate and false alarm rate, we report sensitivity, specificity and their F1. AUC as a decision-threshold-free metric is also reported. 
Authentic images per testset 
are only used for image-level evaluation. 

Note that previous works commonly report performance with the decision threshold selected per testset \cite{2020GSR,2020SPAN,2021CAT}, allowing one to compare models under their optimal conditions. However, this setting leads to overly optimistic performance estimates, as in practice, a model's decision threshold (or its operating point) has to be pre-specified and fixed. Towards real-world evaluation, for both pixel-level and image-level F1 computation, we propose to use a default threshold of $0.5$, unless otherwise stated.


The overall performance is measured by Com-F1, defined as the harmonic mean of pixel-level and image-level F1.  Com-F1 is sensitive to the lowest value of pixel-F1 and
image-F1. In particular, it scores 0 when either pixel-F1 or image-F1 is 0, which does not hold for the arithmetic mean.

\lxr{For a more complete comparison, we additionally report accuracy per test set, \ie the percentage of correctly classified samples in a test set. Pixel-level / image-level accuracy is obtained by treating every pixel / image as a sample. Note that accuracy is not a reliable metric when the class distributions are highly imbalanced.  So we report MCC (Matthews Correlation Coefficients) \cite{mcc2000}, a more balanced measure of a classifier's ability on both classes.}



\textbf{Implementation}. Our models are implemented in PyTorch and trained on an NVIDIA Tesla V100 GPU. The input size is $512\times512$. The two ResNet-50 used in ESB and NSB are initialized with ImageNet-pretrained counterparts. We use an Adam \cite{2014Adam} optimizer with a learning rate periodically decays from $10^{-4}$ to $10^{-7}$. 
For the two hyper-parameters in the combined loss, we empirically set $\alpha=0.16$ and $\beta=0.04$, 
see a parameter sensitivity analysis in the online supplementary material\footnote{\url{https://tinyurl.com/mvssnet-extra}}. As for ConvGeM, the initial value of $p$ in the GeM used in the decayed skip connection is set to $10$ so as to get a similar effect of GMP, while $p$ of the GeM in the Conv branch is initialized to be $3$ to make this GeM more close to global mean pooling. The weight $\lambda$ in Eq. \ref{eq:convgem} is decayed nonlinearly \wrt the number of training epochs $e$ as $\lambda = 0.9975^{(e\cdot e)}, e=1, 2, \ldots$.
An early stop occurs once the loss on a held-out validation set 
from DEFACTO does not decrease in 10 consecutive epochs. As such, the number of training epochs depends on the training data in use: training on CASIAv2 takes 16 epochs, meaning $\lambda$ of 0.527 in the inference mode, while training on DEF-84k requires a larger number of 30 epochs, resulting in a smaller $\lambda$ of 0.105.

We apply regular data augmentation for training, including flipping, blurring, compression and naive manipulations either by cropping and pasting a squared area or using built-in OpenCV inpainting functions \cite{cv2inp1,cv2inp2}. 



%% file: table/table-datasets.tex
\begin{table}[htbp]
\caption{\textbf{Two training sets and six test sets used in our experiments}. The symbol -- indicates information unavailable.  Copy-move, splicing and inpainting are shortened as \textit{cmpv}, \textit{spli} and \textit{inpa}, respectively.  DEF-84k and DEF-12k are used for training and test in the ablation study (Section \ref{ssec:ablation}), while for the SOTA comparison (Section \ref{ssec:eval-sota}) we train on CASIAv2 and evaluate on all test sets. 
}
\renewcommand{\arraystretch}{1.1}
\begin{center}
\scalebox{1}{
\begin{tabular}{@{}lrrrrr@{}}
\toprule
\textbf{Dataset} &
  \multicolumn{1}{r}{\textbf{Negative}} &
  \multicolumn{1}{r}{\textbf{Positive}} &
  \multicolumn{1}{r}{\textbf{cpmv}} &
  \multicolumn{1}{r}{\textbf{spli}} &
  \multicolumn{1}{r}{\textbf{inpa}} \\ \midrule
\multicolumn{1}{@{}l}{\textit{{Training}}}  & \multicolumn{5}{l}{}                      \\ 
DEF-84k\cite{Mahfoudi2019DEFACTO}       & 20,000 & 64,417 & 12,777 & 34,133 & 17,507 \\
CASIAv2\cite{casiav2}           & 7,491  & 5,063  & 3,235  & 1,828  & 0      \\ \midrule
\multicolumn{1}{@{}l}{\textit{{Testing}}}     & \multicolumn{5}{l}{}                      \\ 
COVER\cite{2016COVERAGE}             & 100    & 100    & 100    & 0      & 0      \\ 
Columbia\cite{HsuColumbia}          & 183    & 180    & 0      & 180    & 0      \\ 
NIST16\cite{NIST}            & 0      & 564    & 68     & 288    & 208    \\ 
CASIAv1+\cite{casiav1}           & 800    & 920    & 459    & 461    & 0      \\ 
IMD\cite{2020IMD2020}    & 414  & 2,010  & --  & --  & --  \\ 
DEF-12k\cite{Mahfoudi2019DEFACTO}       & 6,000  & 6,000  & 2,000  & 2,000  & 2,000  \\
\bottomrule
\end{tabular}%
}
\end{center}

\label{table:dataset}
\end{table}

%% file: exp-ablation.tex
To reveal the influence of the individual components, we evaluate the proposed model in varied setups with the components added progressively. All results reported in this section are obtained with DEF-84k as the training set and DEF-12k as the test set.

\subsubsection{\lxr{On Trainable Components}}\label{sssec:eval-trainable}

\textbf{Influence of the semantic segmentation backbone}.  We depart from FCN-16 without multi-view multi-scale supervision. Recall that we use a DA module for branch fusion. So for a fair comparison, we adopt FCN-16 with DA, making it essentially an implementation of DANet~\cite{danet}. Such an improved FCN-16 scores better than its more advanced counterparts, \eg UNet~\cite{unet}, DeepLabv3~\cite{deeplabv3} and DeepLabv3+~\cite{deeplab}, see Table \ref{table:backbone}. The result confirms our conjecture in Section \ref{sec:introduction} that the state-of-the-art semantic segmentation networks are indeed suboptimal for manipulation detection. 
The competitive baseline (FCN-16 with DA) is referred to as \emph{Seg} (Setup \#0) in Table \ref{table:ablation}.
\input{table/table-backbone}

\input{table/table-ablation}

\begin{figure*}[htpb]
    \begin{center}
    \includegraphics[width=1.9\columnwidth]{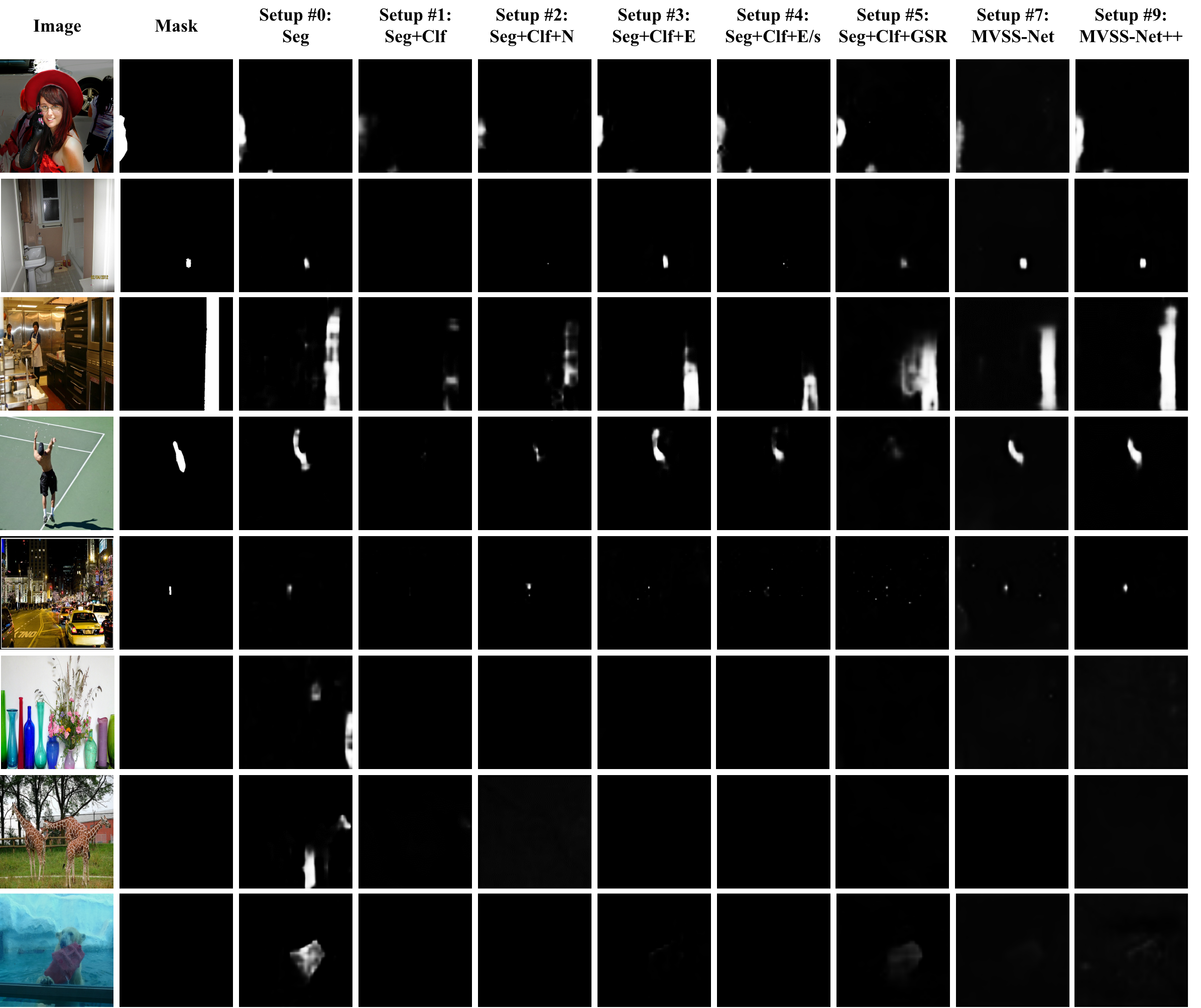}
    \end{center}
    \caption{\textbf{Visualizing pixel-level manipulation detection results of the proposed model in varied setups}. Data source: DEFACTO \cite{Mahfoudi2019DEFACTO}. The test images in the last three rows are authentic. Setups with multi-scale supervision (Seg+Clf and afterwards) improves the detection specificity, yet at the cost of the detection sensitivity, which has to be brought back by multi-view feature learning. Among all the setups, MVSS-Net++ strikes the best balance between the detection sensitivity and specificity.}
    \label{fig:ab_example}
\end{figure*}

\textbf{Influence of the image classification loss}. Comparing \emph{Seg+Clf} and \emph{Seg}, we see a clear increase in specificity and a clear drop in sensitivity, suggesting that adding $loss_{clf}$ makes the model more conservative for reporting manipulation. This change is not only confirmed by lower Pixel-F1, but is also  observed in the fourth column of Fig. \ref{fig:ab_example}, showing that manipulated areas predicted by \emph{Seg+Clf} are much reduced.


\textbf{Influence of NSB}. Since \emph{Seg+Clf+Noise} is obtained by adding NSB into \emph{Seg+Clf}, its better performance verifies the effectiveness of NSB for improving manipulation detection.


\textbf{Influence of ESB}. The better performance of \emph{Seg+Clf+Edge} against \emph{Seg+Clf} justifies the effectiveness of ESB. 



\textbf{ESB versus GSR-Net}. \emph{Seg+Clf+GSR} is obtained by replacing our ESB with the edge branch of GSR-Net \cite{2020GSR}. The overall performance of \emph{Seg+Clf+GSR} is lower than \emph{Seg+Clf+Edge}. Moreover, there is a larger performance gap on \emph{cmpv} (ESB of 0.405 versus GSR-Net of 0.363). The results clearly demonstrate the superiority of the proposed ESB over the prior art.


\textbf{Influence of two branch fusion}. The full setup, with ESB and NSB fused by dual attention, performs the best, showing the complementarity of the individual components. To further justify the necessity of our dual attention based fusion, we make an alternative solution which ensembles \emph{Seg+Clf+Noise} and \emph{Seg+Clf+Edge} by model averaging, refereed to as \emph{Ensemble(N,E)}. Comparing Setup \#6 and \#7 in Table \ref{table:ablation}, we see that MVSS-Net is better than \emph{Ensemble(N,E)}, showing the advantage of our fusion method. 
Comparison to fusion by bilinear pooling as used previously \cite{2018rgbn} is provided in the appendix.

\textbf{Influence of $\mathcal{G}_{clf}$}.
We compare three different implementations of $\mathcal{G}_{clf}$, \ie GMP, GeM and the proposed ConvGeM, with their performance shown in the last three rows of Table \ref{table:ablation}. Compared with GMP, GeM obtains a higher pixel-level F1, indicating a more effective usage of the image-scale supervision for improving the segmentation network. However, GeM averages the responses over pixels, albeit in a nonlinear manner, making it less sensitive, and consequently resulting in a sharp drop in image-level detection sensitivity (from $79.7$ to $63.1$). Its gain on the pixel-level task and its loss on image-level task cancel out each other, making Com-F1 mostly unchanged compared to GMP. By contrast, ConvGeM strikes the best balance between the two tasks, improving Com-F1 from $64.3$ to $66.3$.

\lxr{\textbf{Edge segmentation versus bounding box regression}.
An alternative strategy for learning boundary artifacts around manipulated regions is to treat tampering localization as a bounding-box (bbox) regression task, see \cite{2018rgbn}. To compare with this alternative, we replace the edge segmentation head, \ie the sigmoid layer in Fig. \ref{fig:model}, by the object detection head of CenterNet \cite{2019centernet}. CenterNet performs anchor-free object detection with a two-branch head, where one branch produces a probabilistic map of each pixel being the center of an object, while the other branch is responsible for predicting object sizes. In our context, the region of an object is defined as the minimum bbox that encloses all pixels in a given tempering area. The performance of MVSS-Net trained with the object detection loss is shown in the second last row of Table \ref{table:ablation} (Setup \#9.1). Its relatively lower scores than Setup \#9 in terms of both pixel-level and image-level manipulation detection suggest that the edge segmentation loss is more suited for learning boundary artifact features.
}
\subsubsection{\lxr{On Non-trainable Blocks}} \label{sssec:eval-nontrainable}
\lxr{\textbf{Influence of Sobel on ESB}}. 
\emph{Seg+Clf+Edge/s} is obtained by removing the Sobel operation from \emph{Seg+Clf+Edge}, so its performance degeneration in particular on copy-move detection (from 0.405 to 0.382, \textit{cmpv} in Table \ref{table:ablation}) indicates the necessity of Sobel.

\lxr{It is worth mentioning that the benefit of Sobel for tampering edge detection is concluded on the basis of the ResNet-based network architecture. Compared to ResNet, big vision models with billions of trainable parameters have shown superior performance in natural image classification \cite{vmoe21}. Replacing the ResNet blocks used in MVSS-Net with their counterparts from the big models is likely to boost the performance of the current task. Whether the non-trainable Sobel is beneficial for such more sophisticated network architectures requires future investigation.} 

\lxr{\textbf{Enhancing NSB using non-trainable blocks?}
Inspired by the benefit of Sobel to ESB, we attempt to enhance NSB by using non-trainable blocks to progressively extract noise-related artifacts from the output of each ResNet block in NSB. In particular, we use a median filtering residual (MFR) block after each ResNet block. An MFR processes an input feature map by subtracting the median-filtered feature map from the input, and thus acts as a high-pass filter. The output of each MFR is incrementally aggregated in a shallow-to-deep manner similar to the ESB logic in Fig. \ref{fig:model}. The output of the last MFR is added to the output of the last ResNet block, before the DA module. We refer to the appendix for more details. The performance of NSB with MRF is shown in Setup \#9.2 in Table \ref{table:ablation}. \dcb{Compared} with Setup \#9 (NSB w/o MFR), the higher sensitivity (81.3 \emph{vs} 74.8) and lower specificity (79.9 \emph{vs} 85.7) suggest that high-frequency noise patterns are sensitive to manipulation, but not sufficiently specific, resulting in more pixel-level false alarms. Therefore, the MFR benefit to image-level manipulation detection is obtained at the cost of performance degradation in pixel-level manipulation detection. }


\subsubsection{Qualitative Visualization}

Fig. \ref{fig:ab_example} shows 
some qualitative results of pixel-level manipulation detection. From the left to right, the results demonstrate how the propose model in varied setups  strikes a balance between the detection sensitivity and specificity. 

So far, our evaluation is performed on homologous training and test data. Next, we evaluate the generalization ability of our models in a cross-dataset setting, with CASIAv2 as a common training set and COVER, Columbia, NIST16, CASIAv1+, IMD and DEF-12k as the test sets.

%% file: table/table-backbone.tex
\begin{table}[htbp]
\centering
\renewcommand{\arraystretch}{1.1}
\footnotesize
\caption{\textbf{Performance of different semantic segmentation backbones},  trained with the segmentation loss only. F1 scores are in percentage.}
\scalebox{1}{
\begin{tabular}{lrrrr}
\toprule
\multicolumn{1}{l}{\textbf{Backbone}} & \multicolumn{1}{c}{Pixel-F1} & \multicolumn{1}{c}{Image-F1} & \multicolumn{1}{c}{Image-AUC} & \multicolumn{1}{c}{Com-F1} \\ \midrule

U-Net            & 13.2 & 51.7 & 0.540 & 21.0 \\
DeepLabV3        & 24.9 & 52.6 & 0.645 & 33.8 \\
DeepLabV3+       & 27.9 & 50.9 & 0.651 & 36.0 \\
FCN-16          & 33.7 & 69.9 & 0.774 & 45.5 \\
FCN-16 with DA & \incolor{54.6} & \incolor{70.9} & \incolor{0.840}& \incolor{61.7} \\
\bottomrule
\end{tabular}%
}\\

\label{table:backbone}
\end{table}

%% file: table/table-ablation.tex
\begin{table*}[htbp]
\caption{\textbf{Ablation study of \model}. Training: DEF-84k. Test: DEF-12k. F1, Sensitivity (\textit{Sen.}) and Specificity (\textit{Spe.}) scores are in percentage. Best number per column is highlighted in \textbf{bold}. The steadily improved performance justifies the necessity of the individual components used in \model~(Setup \#7) and \modelplus~(Setup \#9).}
\begin{center}
\renewcommand{\arraystretch}{1.1}
\footnotesize
\scalebox{0.99}{

\begin{tabular}{lccccp{0.8cm}<{\centering}p{0.8cm}<{\centering}p{0.8cm}<{\centering}p{0.8cm}<{\centering}cp{0.7cm}<{\centering}p{0.7cm}<{\centering}p{0.7cm}<{\centering}p{0.7cm}<{\centering}c}
\toprule
\multirow{2}{*}{\textbf{Setup}} & &
  \multicolumn{2}{c}{\textbf{Component}} & &
  \multicolumn{4}{c}{\textbf{Pixel-level manipulation detection (F1)}} &&
  \multicolumn{4}{c}{\textbf{Image-level manipulation detection}} &
  \multicolumn{1}{c}{\multirow{2}{*}{\textbf{Com-F1}}} \\ \cline{3-4}\cline{6-9}\cline{11-14}
 &
  &
  \textit{ESB} &
  \textit{NSB} &&
  \textit{cpmv.} &
  \textit{spli.} &
  \textit{inpa.} &
  \textit{MEAN} & &
  \textit{AUC} &
  \textit{Sen.} &
  \textit{Spe.} &
  \textit{F1} &
  \multicolumn{1}{c}{} \\ \midrule
\multicolumn{12}{l}{\emph{Conference version}\cite{iccv21-mvssnet}, with GMP as $\mathcal{G}_{clf}$ } \\
~0:~Seg   && 
-- & --           && 45.3 & 72.2 & 46.3 & 54.6 && 0.840 & \incolor{82.7} & 62.0 & 70.9 & 61.7 \\ 
~1:~Seg+Clf  && 
-- & --           && 34.1 & 67.3 & 37.6 & 46.3 && 0.858 & 76.8 & 77.8 & 77.3 & 57.9 \\ 
~2:~Seg+Clf+Noise   && 
-- & +           && 39.3 & 70.6 & 42.6 & 50.8 && 0.871 & 76.3 & 82.1 & 79.1 & 61.9 \\ 
~3:~Seg+Clf+Edge   &&
+ & --           && 40.5 & 71.5 & 43.5 & 51.8 && 0.870 & 77.3 & 81.1 & 79.2 & 62.6 \\ 
~4:~Seg+Clf+Edge/s  &&
w/o sobel & --   && 38.2 & 71.0 & 42.2 & 50.5 && 0.869 & 79.2 & 78.9 & 79.0 & 61.6 \\ 
~5:~Seg+Clf+GSR  &&
GSR-Net   & --   && 36.3 & 71.4 & 42.1 & 49.9 && 0.864 & 81.3 & 77.9 & 79.6 & 61.3 \\ 
~6:~Ensemble(\#2, \#3)  &&
+ & +           && 38.4 & 70.8 & 43.7 & 51.0 && 0.878 & 73.1 & 87.6 & 79.7 & 62.2 \\

~7: Seg+Clf+Noise+Edge  &&
+ & +           && 44.6 & 71.4 & 45.5 & 53.8 && 0.886 & 79.7 & 80.2 & 79.9 & 64.3 \\ \midrule

\multicolumn{12}{l}{\emph{Journal extension, built on top of Setup \#7}} \\ 
~8: GeM as $\mathcal{G}_{clf}$  && 
+ & +           && 48.0 & 73.5 & 47.0 & 56.2 && 0.871 & 63.1 & \incolor{93.0} & 75.2 & 64.3\\
  
~9: ConvGeM as $\mathcal{G}_{clf}$  &&
+ & + && \incolor{48.3} & 72.8 & \incolor{49.0} & \incolor{56.7} && 0.879 & 74.8 & 85.7 & 79.9 & \incolor{66.3} \\ 

~\lxr{9.1: Edge $\rightarrow$ BBox}  &&
+ & + && 47.7 &	\incolor{73.9} &	47.1 &	56.2 &&	0.884 &	70.6 &	89.4 &	78.9 &	65.7 \\ 

~\lxr{9.2: NSB with MFR}  &&
+ & + && 45.5 &	72.4 &	46.0 &	54.6 &&	\incolor{0.894} &	81.3 &	79.9 & \incolor{80.6} &	65.1 \\

\bottomrule
\end{tabular}%
}

\end{center}

\label{table:ablation}
\end{table*}

%% file: exp-sota.tex

\input{exp-baselines}

\input{table/table-pixel-public}

\input{table/table-img-public}

\subsubsection{Pixel-Level Manipulation Detection}

%
Table \ref{table:pix-f1} shows the pixel-level detection performance of the varied models. \modelplus~is the best in terms of overall performance.
We attribute the better performance of ManTra-Net on DEF-12k to its large-scale training data, which was also originated from MS-COCO as DE-12k. The top performer on NIST is H-LSTM, the training data of which contains around 70\% of NIST. Compared with baselines trained on the same CASIAv2, \ie MFCN, RGB-N, CR-CNN and GSR-Net, \modelplus ~surpasses them on almost all testsets. Its superior performance in this cross-dataset setting justifies its better generalization ability.   

Comparing the left part of Table \ref{table:pix-f1}, which shows the models' performance in their optimal conditions, and the right part of the table, which shows the counterpart performance in a real scenario, a clear gap exists. For the best baseline, \ie SPAN, its pixel-level F1 drops from $68.8$ to $21.4$. As for \modelplus, its F1 drops from $73.2$ to $38.7$. The result shows the challenging nature of the task and the necessity of the proposed evaluation protocol for fairly assessing the technical progress towards real deployment.
  





\subsubsection{Image-Level Manipulation Detection}

\begin{figure}[htbp]
    \centering
    \includegraphics[width=0.9\linewidth]{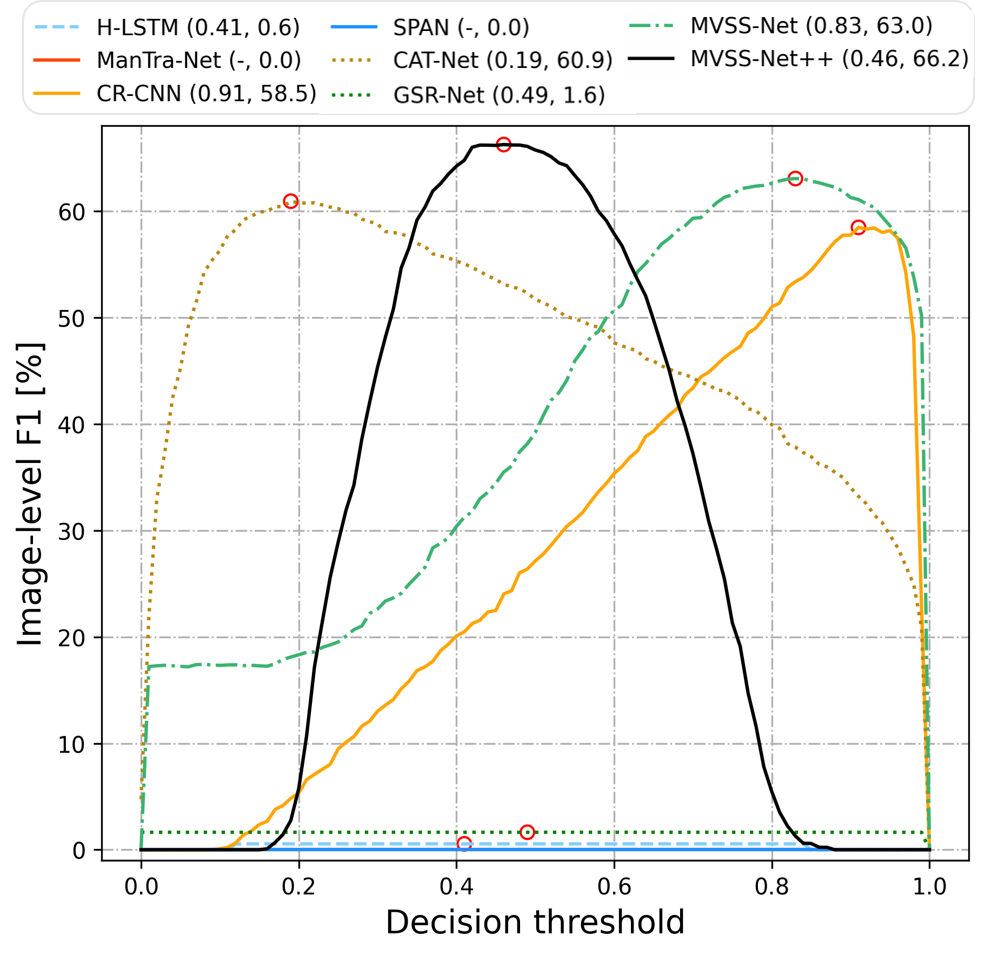}
    \caption{\textbf{Performance curves \wrt the decision threshold}. Larger threshold means lower sensitivity and higher specificity. Per model, its image-level F1 score given a specific threshold is obtained by averaging the F1 scores over the five testsets, \ie Columbia, CASIAv1+, COVER, DEF-12k and IMD. The two numbers following each model are its optimal threshold and the corresponding F1, as visualized in red circles.}
    \label{fig:theta-f1}
\end{figure}

 Table \ref{table:perf-img} shows the image-level performance of the different models, all using the default decision threshold of 0.5. \modelplus~is again the top performer. With multi-scale supervision, the \model ~series are able to learn from the authentic and obtains higher specificity, and thus lower false alarm rate, on most test sets.
 Our models also have competitive AUC scores, meaning they are better than the baselines on a wide range of operating points. 
 Fig. \ref{fig:theta-f1} shows the performance curves of the individual models \wrt the decision threshold. The peak performance of \modelplus ~is obtained at the decision value of $0.46$, much closer to $0.5$ than its counterparts in the other models. This result again suggests the better generalization ability of our model.
 

The overall performance of both pixel-level and image-level manipulation detection is provided in Table \ref{table:perf-img-comf1}.

\input{exp-robustness}

\subsubsection{Efficiency Test} 
%
We measure the inference efficiency in terms of frames per second (FPS), tested on two GPU cards, \ie NVIDIA Tesla V100 (32GB GPU memory footprint) and GeForce RTX2080ti (12GB GPU memory footprint), respectively. Depending on the card in use, the \model~series run at $16$ to $20$ FPS, see Table \ref{table:fps}. The relatively high FPS as compared to the other publicly available models permits the \model~series for real-time application.  
\input{table/table-fps}

\subsubsection{Failure Case Analysis} \label{sssec:limitations}

\input{exp-limitations}

%% file: exp-baselines.tex

\subsubsection{Baselines}

For a fair and reproducible comparison, we have to be selective, choosing the state-of-the-art that meets one of the following three criteria: 1) pre-trained models  released by paper authors, 2) source code  publicly available, and 3) following a common evaluation protocol where CASIAv2 is used for training and other public datasets are used for testing. Accordingly, we compile a list of nine published baselines as follows:\\
$\bullet$ Models available: 
H-LSTM~\cite{H-LSTM} (pre-trained on a homemade dataset of 65k manipulated images and finetuned on NIST16 and IEEE Forensics
Challenge data\footnote{\url{https://github.com/jawadbappy/forgery_localization_HLED}}),  
ManTra-Net~\cite{mantranet} (trained on a private set of millions of manipulated images\footnote{\url{https://github.com/ISICV/ManTraNet}}),
HP-FCN \cite{HPFCN} (trained on a private set of inpainted images\footnote{\url{https://github.com/lihaod/Deep_inpainting_localization}}), 
CR-CNN~\cite{2020Constrained} (trained on CASIAv2\footnote{\url{https://github.com/HuizhouLi/Constrained-R-CNN}}),  SPAN~\cite{2020SPAN} (trained on the same data as ManTra-Net and finetuned on CASIAv2\footnote{\url{https://github.com/ZhiHanZ/IRIS0-SPAN}}), and  CAT-Net~\cite{2021CAT} (trained on a joint dataset including CASIAv2, IMD, Fantastic Reality\cite{Fantastic}, self-spliced COCO\footnote{\url{https://github.com/mjkwon2021/CAT-Net}}). We use these six models as is. \\
$\bullet$ Code available: GSR-Net \cite{2020GSR}, which we train using author-provided  code\footnote{\url{https://github.com/pengzhou1108/GSRNet}}. We cite their results where appropriate and use our re-trained model only when necessary. \\
$\bullet$ Same evaluation protocol: MFCN \cite{2017MFCN} and RGB-N \cite{2018rgbn} with numbers quoted from the same team \cite{2020GSR}. 

For a fair comparison, we have re-trained FCN~(Setup\#0 in Table \ref{table:ablation}), \model~(Setup\#7) and \modelplus(Setup\#9) from scratch on CASIAv2. As the previous works seldom report their image-level performance, an image classification head is naturally missing in their implementations. In order to obtain image-level predictions of the baselines yet with no need of hacking into their models or code, we utilize GMP as having been used in \model.

%% file: table/table-pixel-public.tex
\begin{table*}[htbp]
\caption{\textbf{Performance on pixel-level manipulation detection}. Performance metric: F1 [\%]. Best result per test set is highlighted in bold font. All models are trained on CASIAv2, except for those marked with star (*), the training data of which contains either private (ManTra-Net, SPAN, CAT-Net and HP-FCN) or published but no longer publicly accessible data (H-LSTM).} 
\renewcommand{\arraystretch}{1.1}
\begin{center}
\footnotesize
\setlength{\tabcolsep}{0.7mm}

\scalebox{1}{
\begin{tabular}{lccccccccp{0.14cm}<{\centering}ccccccc}
\toprule
\multirow{2}{*}{\textbf{Method}} && \multicolumn{7}{c}{\textbf{Optimal threshold per model \& testset}} && \multicolumn{7}{c}{\textbf{Fixed threshold (0.5)}}  \\
\cline{3-9}\cline{11-17}
&& \textit{NIST} & \textit{Columbia} & \textit{CASIAv1+}  & \textit{COVER}  & \textit{DEF-12k}  & \textit{IMD} & \textit{MEAN} && \textit{NIST} & \textit{Columbia} & \textit{CASIAv1+} & \textit{COVER}    & \textit{DEF-12k}    & \textit{IMD} & \textit{MEAN}  \\ \hline
MFCN, VCIR17\cite{2017MFCN}&& 
42.2    & 61.2       & 54.1      & n.a.     & n.a.      & n.a.    & n.a.  &&
n.a.     & n.a.    & n.a.  & n.a.    & n.a. & n.a. &n.a.  \\
RGB-N, CVPR18\cite{2018rgbn} &&                  
n.a.     & n.a.        & 40.8      & 37.9    & n.a.      & n.a.    & n.a.  &&
n.a.     & n.a.    & n.a.  & n.a.    & n.a. & n.a. & n.a.   \\
H-LSTM*, TIP19\cite{H-LSTM} &&
46.6  & 	14.2 &	20.9 &	21.3 &	12.5 &	31.0 & 24.4 &&
\incolor{35.4} &	13.0 &	15.4 &	16.3 	& ~~5.9 & 19.5 & 17.6 \\
ManTra-Net*, CVPR19\cite{mantranet} &&
45.5    & 70.9       & 69.2      & 77.2    & \incolor{61.8}     & 70.5 &65.9  && 
~~0.0 & 36.4    & 15.5   & 28.6 & \incolor{15.5}   & 18.7 & 19.1 \\
HP-FCN*, ICCV19\cite{HPFCN} &&
36.0 &	47.1 &	21.4	& 19.9 &	13.6 & 16.9 & 25.8 && 
12.1 &	~~6.7	& 15.4 &	~~0.3	& ~~5.5 & 11.2 & ~~8.5\\
CR-CNN, ICME20\cite{2020Constrained} &&
42.8    & 70.4       & 66.2      & 47.0    & 34.0  &60.0   & 53.4   && 
23.8 & 43.6    & 40.5   & 29.1 & 13.2   & 26.2  &29.4\\
GSR-Net, AAAI20\cite{2020GSR}  &&
45.6    & 62.2       & 57.4      & 48.9    & 37.9     & 68.7  &53.4 &&
28.3 & 61.3    & 38.7   & 28.5 & ~~5.1   & 24.3 &31.0\\
SPAN*, ECCV20\cite{2020SPAN} &&
68.3 &	77.4 & 	68.8 &	71.8 &	57.1 &	69.6 & 68.8 &&
22.1 &	48.7 & 	18.4 &	17.2 &	~~4.8 &	17.0&21.4\\
CAT-Net*, WACV20\cite{2021CAT} &&
59.9 &	\incolor{77.6} &	57.3 &	48.5 &	44.1 &	51.7 & 56.5 && 
17.9 &	55.5 &	13.6 & 	12.9 &	~~4.6 &	~~5.4 & 18.3 \\
\midrule
\multicolumn{15}{l}{\emph{This paper:}}\\
FCN && 
50.7 &	58.6 &	74.2 &	57.3 &	40.1 &	64.5 & 57.6 && 
16.7 &	22.3 &	44.1 &	19.9 &	13.0 &	21.0 & 22.8\\
\model && 
\incolor{73.7}    & 70.3       & 75.3      &  82.4    & 57.2     &  75.7  & 72.4  && 
29.2 & 63.8    & 45.2   & 45.3 & 13.7   & 26.0 & 37.2 \\ 
\modelplus&&
 71.5 & 73.1 &	\incolor{77.1} &	\incolor{83.2} &	55.6 &	\incolor{76.2} & \incolor{72.8} &&
30.4 &	\incolor{66.0} &	\incolor{51.3} &	\incolor{48.2} & 	~~9.5 &	\incolor{27.0} &\incolor{38.7}\\

\bottomrule
\end{tabular}
}
\end{center}

\label{table:pix-f1}
\end{table*}

%% file: table/table-img-public.tex
\begin{table*}[htpb]
\caption{\textbf{Performance on image-level manipulation detection}. Decision threshold: $0.5$. NIST16 is excluded as it has no authentic image. \modelplus~tops the performance with image-level F1 of $68.0$ averaged over the five test sets, followed by CAT-Net ($51.7$) and \model~ ($51.2$).}
\renewcommand{\arraystretch}{1.1}
\begin{center}
\footnotesize
\setlength{\tabcolsep}{1mm}
\scalebox{1}{
\begin{tabular}{lcrrrrcrrrrcrrrrcrrrrcrrrr}
\toprule
\multicolumn{1}{l}{\multirow{2}{*}{\textbf{Method}}} &&
  \multicolumn{4}{c}{\textbf{Columbia}} &&
  \multicolumn{4}{c}{\textbf{CASIAv1+}} &&
  \multicolumn{4}{c}{\textbf{COVER}} &&
  \multicolumn{4}{c}{\textbf{DEF-12k}} &&
  \multicolumn{4}{c}{\textbf{IMD}}\\ \cline{3-6}\cline{8-11}\cline{13-16}\cline{18-21}\cline{23-26}
\multicolumn{1}{c}{} &&
  \textit{AUC} &
  \textit{Sen.} &
  \textit{Spe.} &
  \textit{F1} &&
  \textit{AUC} &
  \textit{Sen.} &
  \textit{Spe.} &
  \textit{F1} &&
  \textit{AUC} &
  \textit{Sen.} &
  \textit{Spe.} &
  \textit{F1} &&
\textit{AUC} &
  \textit{Sen.} &
  \textit{Spe.} &
  \textit{F1} &&
  \textit{AUC} &
  \textit{Sen.} &
  \textit{Spe.} &
  \textit{F1} \\ \midrule
H-LSTM &&
0.506 & 100.0 & 1.1 & 2.2 &&
0.498 & 99.7  & 0.0 & 0.0 &&
0.500 & 100.0 &	0.0 & 0.0 &&
0.499 & 99.9  & 0.1 & 0.2 &&
0.501 & 100.0 & 0.0 & 0.0 \\  
ManrTra-Net&&
0.701 &	100.0 &	0.0 & 0.0 &&
0.500 &	100.0 &	0.0 & 0.0 &&
0.500 &	100.0 &	0.0 & 0.0 &&
0.543 &	100.0 &	0.0 & 0.0 &&
0.500 &	100.0 &	0.0 & 0.0 \\
CR-CNN &&
0.783 &	96.1 &	24.6 &	39.2 &&
0.719 &	93.0 &	13.9 &	24.2 &&
0.566 &	96.7 &	7.0 &	13.1 &&
0.567 &	77.4 &	26.7 &	39.7 &&
0.615 &	92.9 &	12.3 &	21.7 \\
GSR-Net &&
0.502 &	100.0 &	1.1 &	2.2 &&
0.500 &	99.4 &	0.0 &	0.0 &&
0.515 &	100.0 &	0.0 &	0.0 &&
0.456 &	91.4 &	0.1 &	0.2 &&
0.500 &	100.0 &	0.0 &	0.0 \\
SPAN &&
0.500 & 100.0 & 0.0 & 0.0 &&
0.500 & 100.0 & 0.0 & 0.0 &&
0.500 & 100.0 &	0.0 & 0.0 &&
0.500 & 100.0 & 0.0 & 0.0 &&
0.500 & 100.0 & 0.0 & 0.0 \\
CAT-Net &&
0.971 & 87.2 & 96.2 & 91.5 &&
0.647 & 23.9 & 92.1 & 38.0 &&
0.557 & 28.0 &	80.0 & 41.5 &&
0.543 & 34.2 & 72.5 & 46.5 &&
0.586 & 27.5 & 81.6 & 41.1 \\\midrule
FCN &&
0.762 &	95.0 &	32.2 &	48.1 &&
0.770 &	72.8 &	64.3 &	68.3 &&
0.541 &	90.0 &	10.0 &	18.0 &&
0.551 &	71.1 &	33.8 &	45.8 &&
0.502 &	84.6 &	15.5 &  26.2 \\
\model &&
0.980 & 66.9 & 100.0 & 80.2 &&
\incolor{0.937} & 61.5 & 98.8 & \incolor{75.8} &&
\incolor{0.731} & 94.0 &	14.0 & 24.4 &&
\incolor{0.573} & 81.7 & 26.8 & 40.4 &&
0.656 & 91.5 & 22.0 & 35.5 \\
\modelplus&&
\incolor{0.984} & 96.7 & 89.6 & \incolor{93.0} &&
0.862 & 53.6 & 98.4 & 69.4 &&
0.726 & 69.0 &	68.0 & \incolor{68.5} &&
0.531 & 37.3 &	66.6 & \incolor{47.8} &&
\incolor{0.658} & 59.5 & 63.5 & \incolor{61.4} \\ 


\bottomrule
\end{tabular}%
}
\end{center}

\label{table:perf-img}
\end{table*}

\begin{table}[htpb]
\caption{\textbf{Overall performance measured by Com-F1}, the harmonic mean of pixel-level F1 and image-level F1, on five test sets. }
\begin{center}
\renewcommand{\arraystretch}{1.1}
\footnotesize
\setlength{\tabcolsep}{0.8mm}
\scalebox{1}{
\begin{tabular}{@{}lcrrrrrr@{}}
\toprule
\textbf{Method} & &
  \textbf{Columbia} &
  \textbf{CASIAv1+} &
  \textbf{COVER} &
  \textbf{DEF-12k} &
  \textbf{IMD} &
  \textbf{MEAN}\\ \midrule
H-LSTM&	&        3.8 &	0.0 &	0.0 &	0.4 &	0.0 & 0.8 \\
ManrTra-Net& &	0.0 &	0.0 &	0.0 &	0.0 &	0.0 & 0.0\\
CR-CNN&    & 	41.3 &	30.3 &	18.1 &	19.8 &	23.7& 26.6 \\
GSR-Net&    &  	4.2 &	0.0 &	0.0 &	0.4 &	0.0 & 0.9\\
SPAN&	   &     0.0 &	0.0 &	0.0 &	0.0 &	0.0 & 0.0\\
CAT-Net&	&   69.1 &	20.0 &	19.7 &	8.4 &	9.5 & 25.3\\ \midrule
FCN & & 30.5 &	53.6 &
18.9&	20.3 &	23.3 & 29.3\\
\model &	&    71.1 &	56.6 &	31.7 &	\incolor{20.5} &	30.0 & 42.0\\
\modelplus&	&    \incolor{77.2} &	\incolor{59.0} &	\incolor{56.6} &	15.8 &	\incolor{37.5} & \incolor{49.2} \\
\bottomrule
  
\end{tabular}%
}

\end{center}

\label{table:perf-img-comf1}
\end{table}

%% file: exp-robustness.tex


\subsubsection{Robustness Evaluation} 

Following \cite{2018rgbn,mantranet,2020SPAN}, we evaluate the model robustness against two daily image processing operations when images circulate on the Internet, \ie JPEG compression and Gaussian blur. Furthermore, we investigate how the current models react to manipulated images re-captured by screenshot, which to the best of our knowledge has not been done before.

\begin{figure}[htbp]
\begin{center}
\begin{minipage}[t]{0.96\linewidth}
\includegraphics[width=0.96\textwidth]{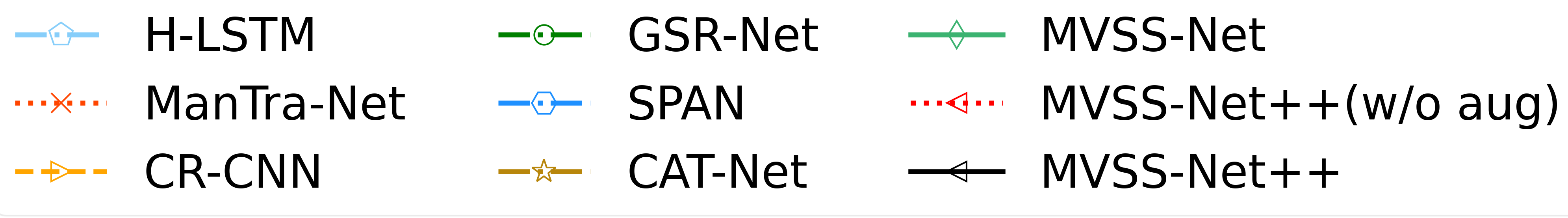}
\end{minipage}%
\\
\subfigure[~]{
\includegraphics[width=0.48\linewidth]{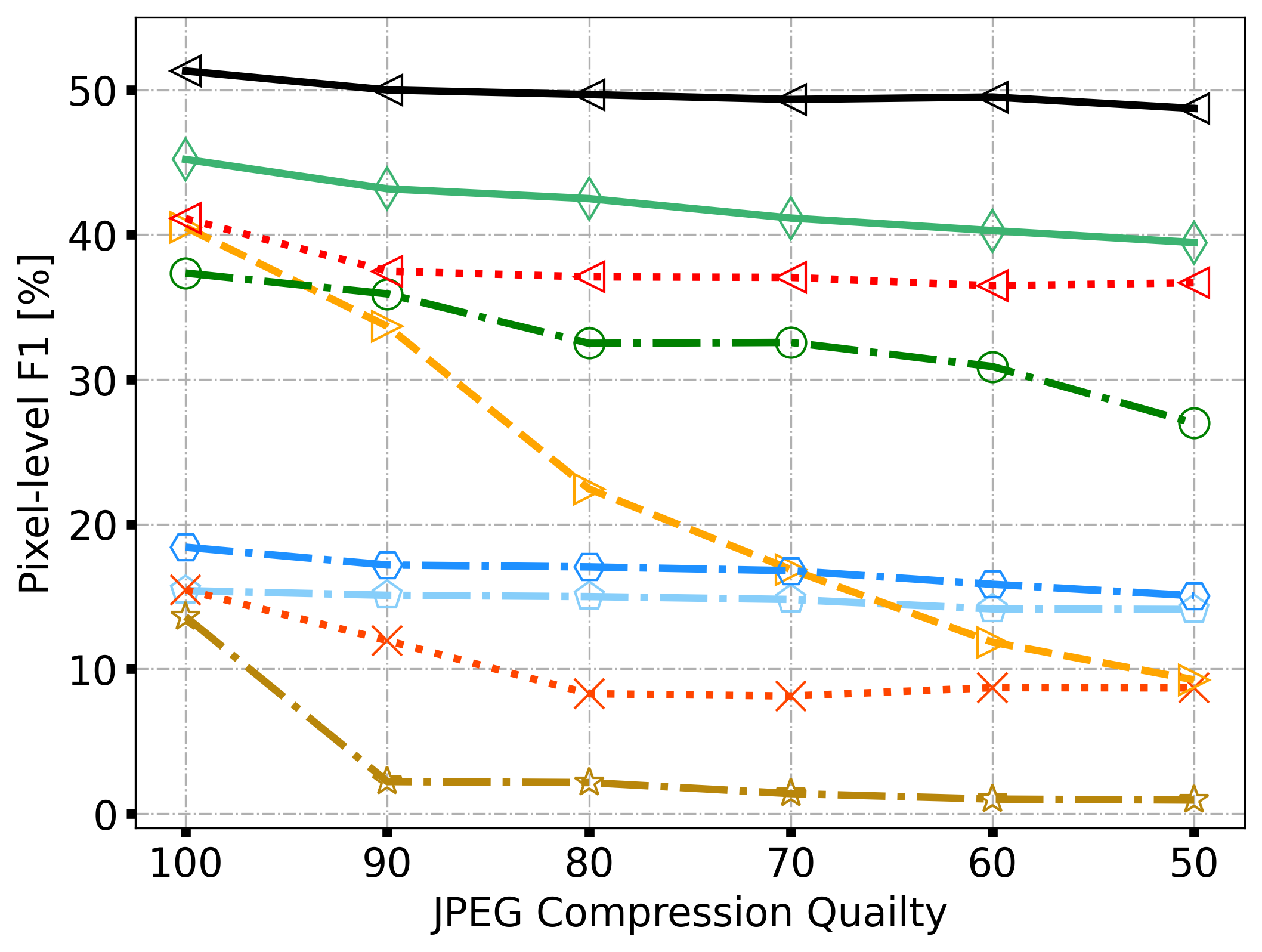}
}%
\subfigure[~]{
\includegraphics[width=0.48\linewidth]{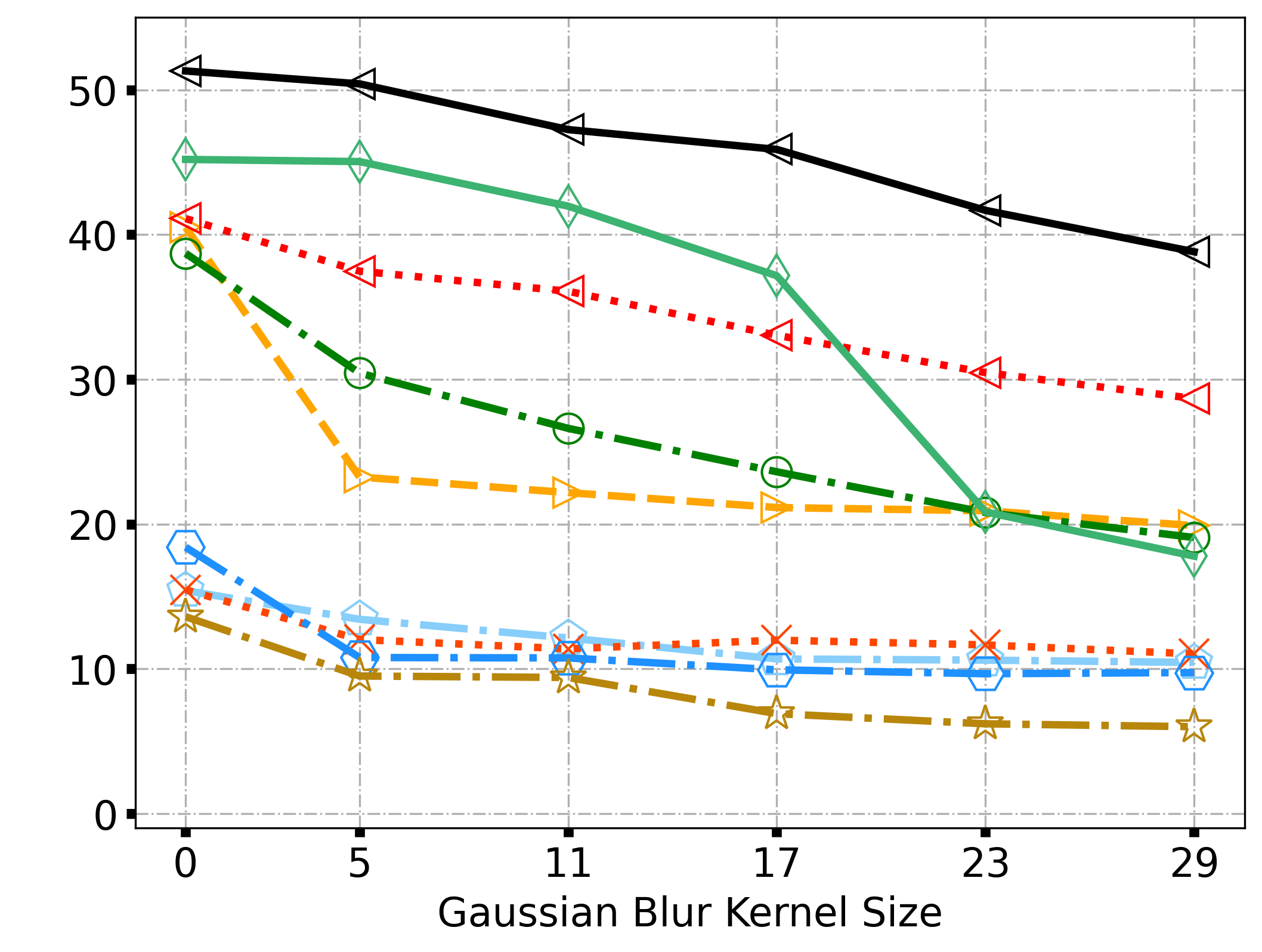}
}%
\\\vspace{.01in}
\subfigure[~]{
\includegraphics[width=0.48\linewidth]{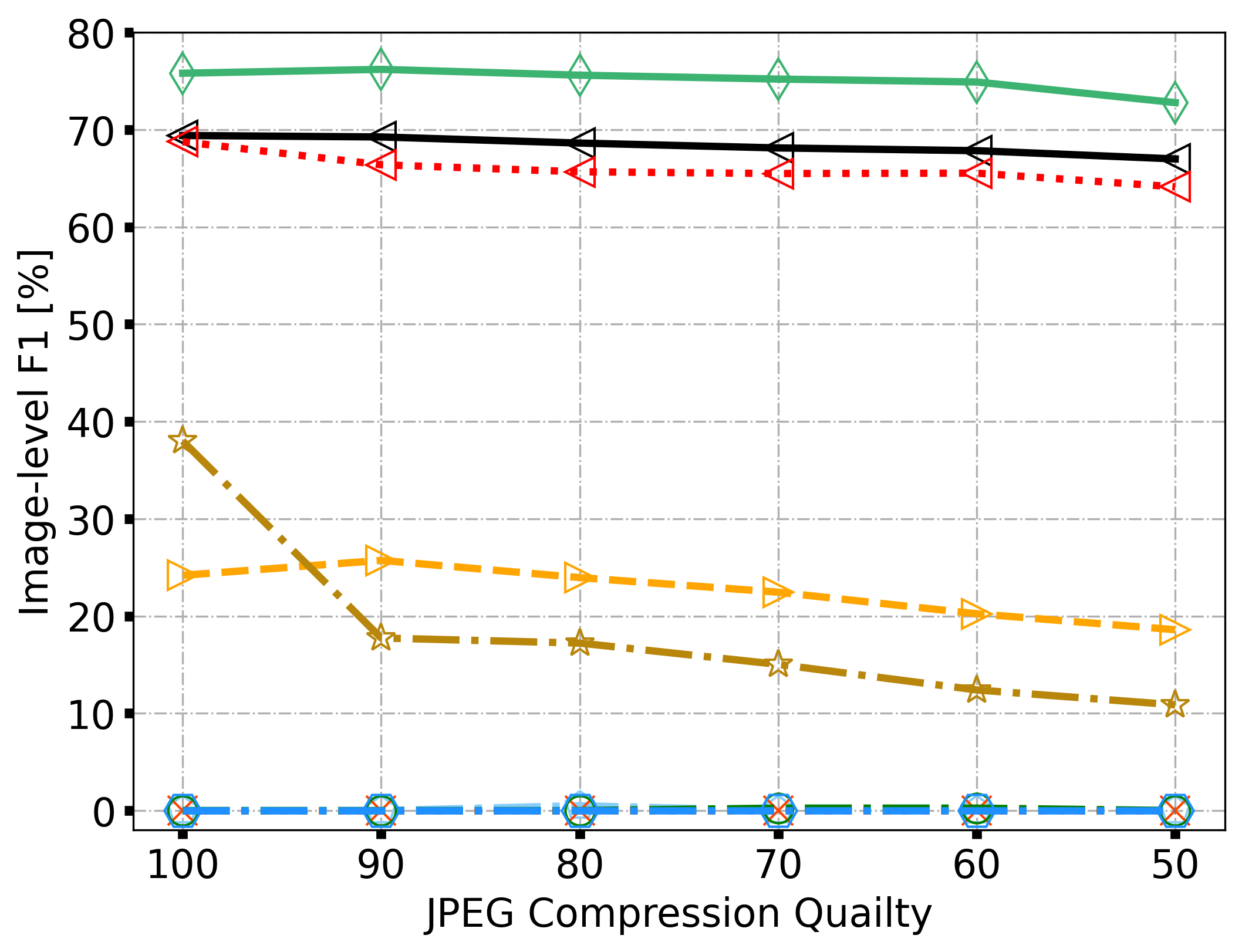}
}%
\subfigure[~]{
\includegraphics[width=0.48\linewidth]{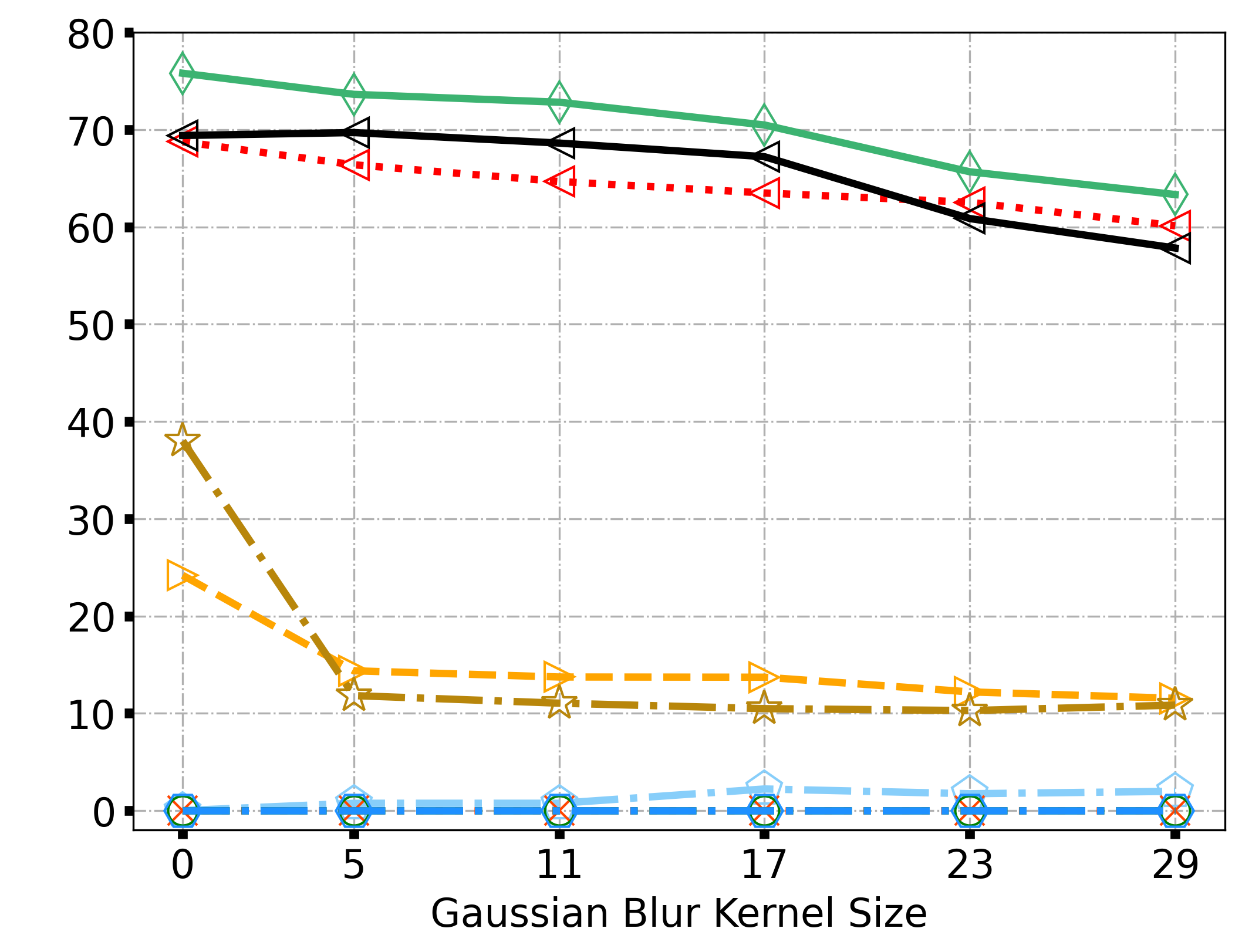}
}%
\\\vspace{.01in}
\subfigure[~]{
\includegraphics[width=0.48\linewidth]{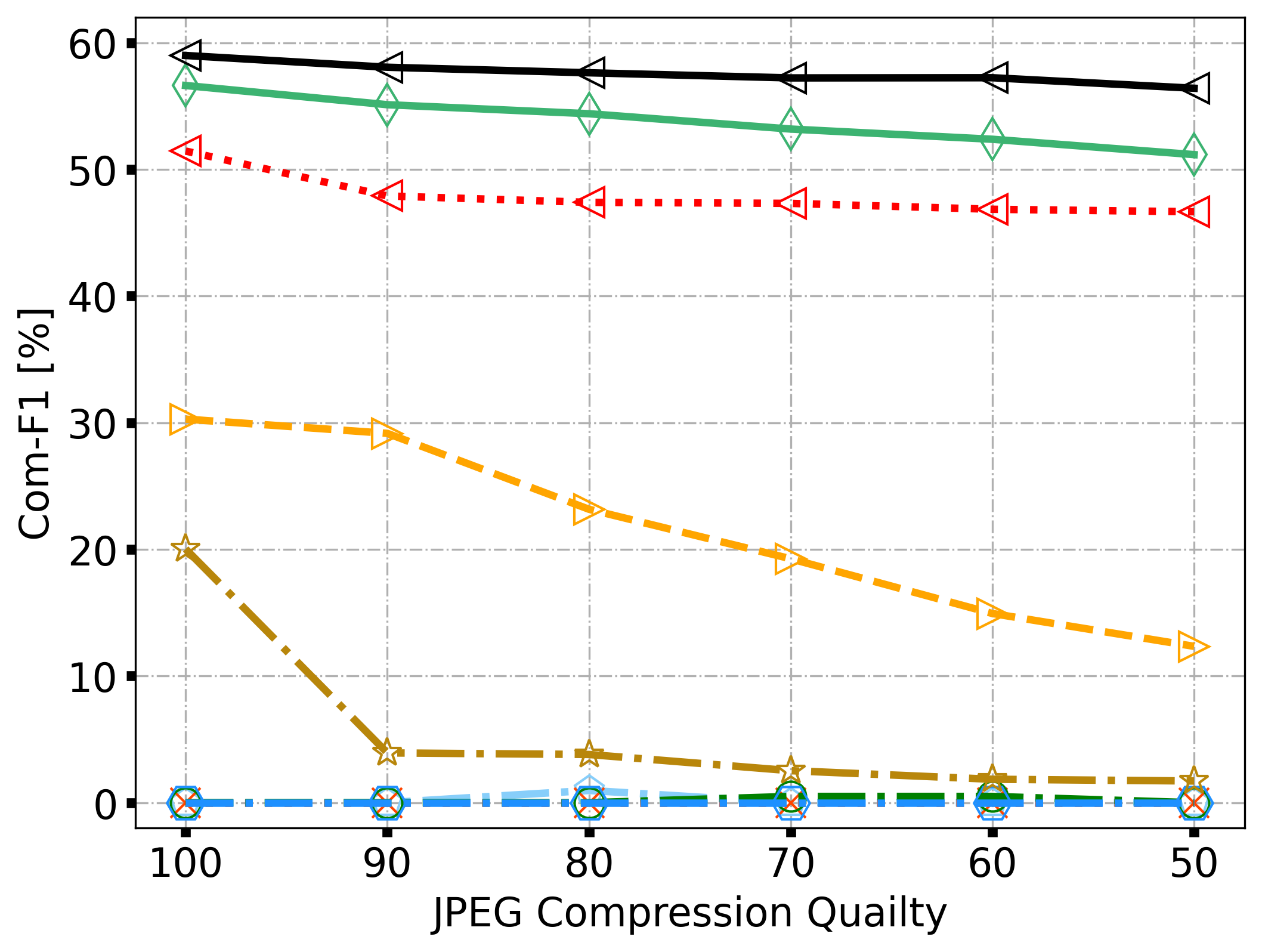}
}%
\subfigure[~]{
\includegraphics[width=0.48\linewidth]{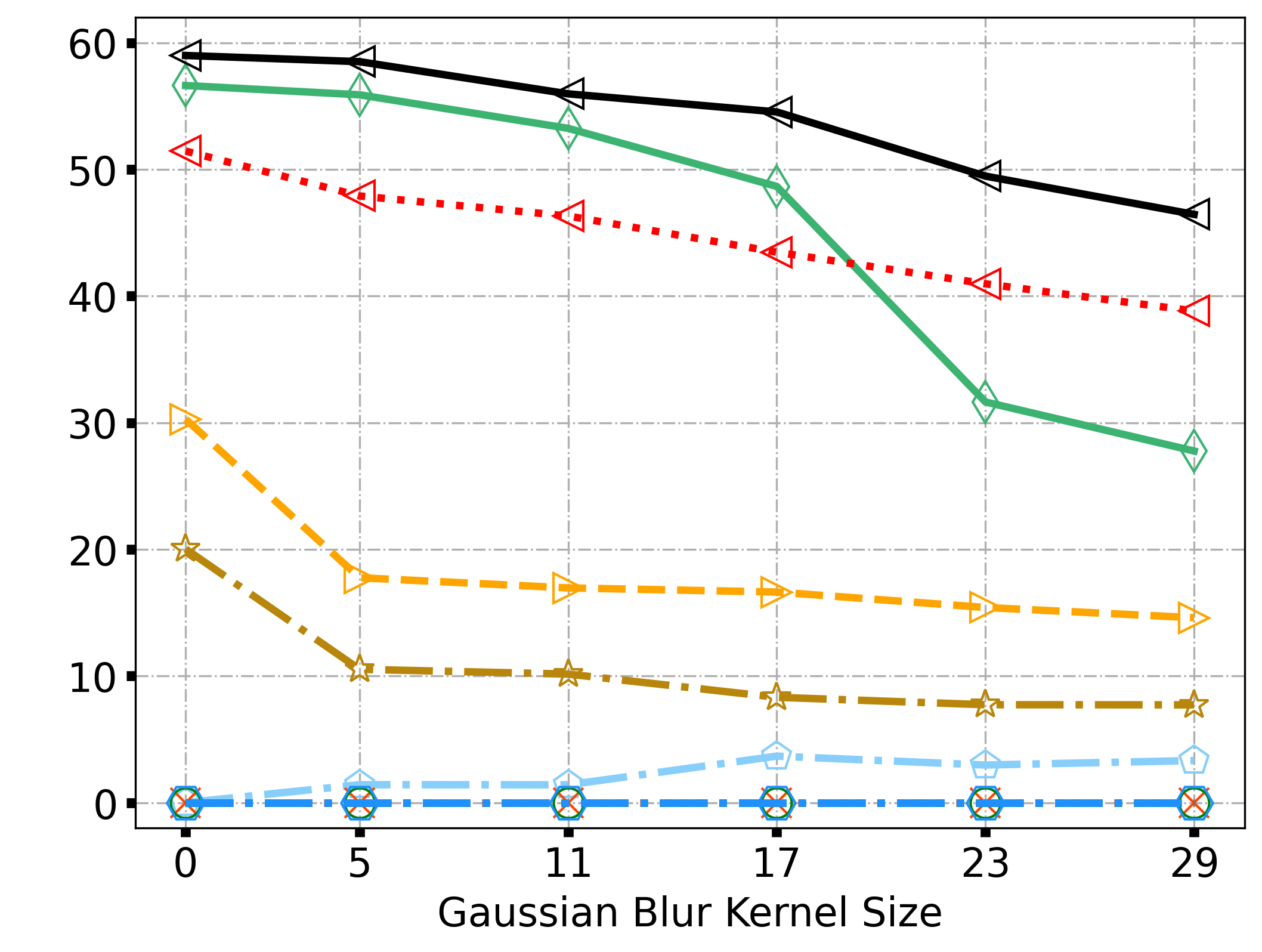}
}%

\end{center}
\caption{\textbf{Robustness evaluation against two image processing techniques, \ie JPEG compression and Gaussian Blurs}. Test set: CASIAv1+. \modelplus~(w/o aug) indicates training with JPEG compression and Gaussian blur excluded from data augmentation.  The proposed models are more robust than the baselines.}

\label{fig:casia_robust}
\end{figure}

Comparing the two operations, Gaussian blur affects the detection performance more severely, in particular when a larger kernel size of $17\times17$ and above are used, see Fig. \ref{fig:casia_robust}. While such a larger kernel effectively erase manipulation traces, it also noticeably decreases the readability of the pictorial content (data not shown). Both \model ~and \modelplus ~exhibit better robustness than the baselines. 
According to their original papers, ManTra-Net and SPAN used a wide range of data augmentations including compression, while CR-CNN, GSR-Net and CAT-Net did not use such data augmentation. So for a more fair comparison, we also train \modelplus~ with compression and blur excluded from data augmentation. The re-trained model, denoted as \modelplus~(w/o aug), remains more robust than the baselines.

The screenshot oriented evaluation is conducted as follows. A subset of 100 manipulated images are randomly selected from CASIAv1+. Each image is then manually re-captured on a Windows10 laptop with a screen resolution of $1920 \times 1080$. Three commercial screenshot tools are used, including Microsoft \emph{Snip\&Sketch}\footnote{\url{https://www.microsoft.com/en-us/p/snip-sketch}}, Google \emph{Chrome DevTools}\footnote{\url{https://developer.chrome.com/blog/new-in-devtools-74/\#screenshot}}, and \textit{Snipaste}\footnote{\url{https://www.snipaste.com/}}. Per image and tool, the result image is saved in jpg (with a quality level of 90\%) and png formats, respectively, except for Chrome which supports png only. Varying the configuration of the screen tool and the image format results into five variants of the test set, \ie Snip\&Sketch (png), Snip\&Sketch (jpg), Chrome (png),  Snipaste (png) and Snipaste (jpg).

\begin{figure}[htbp]
    \centering
    \includegraphics[width=\linewidth]{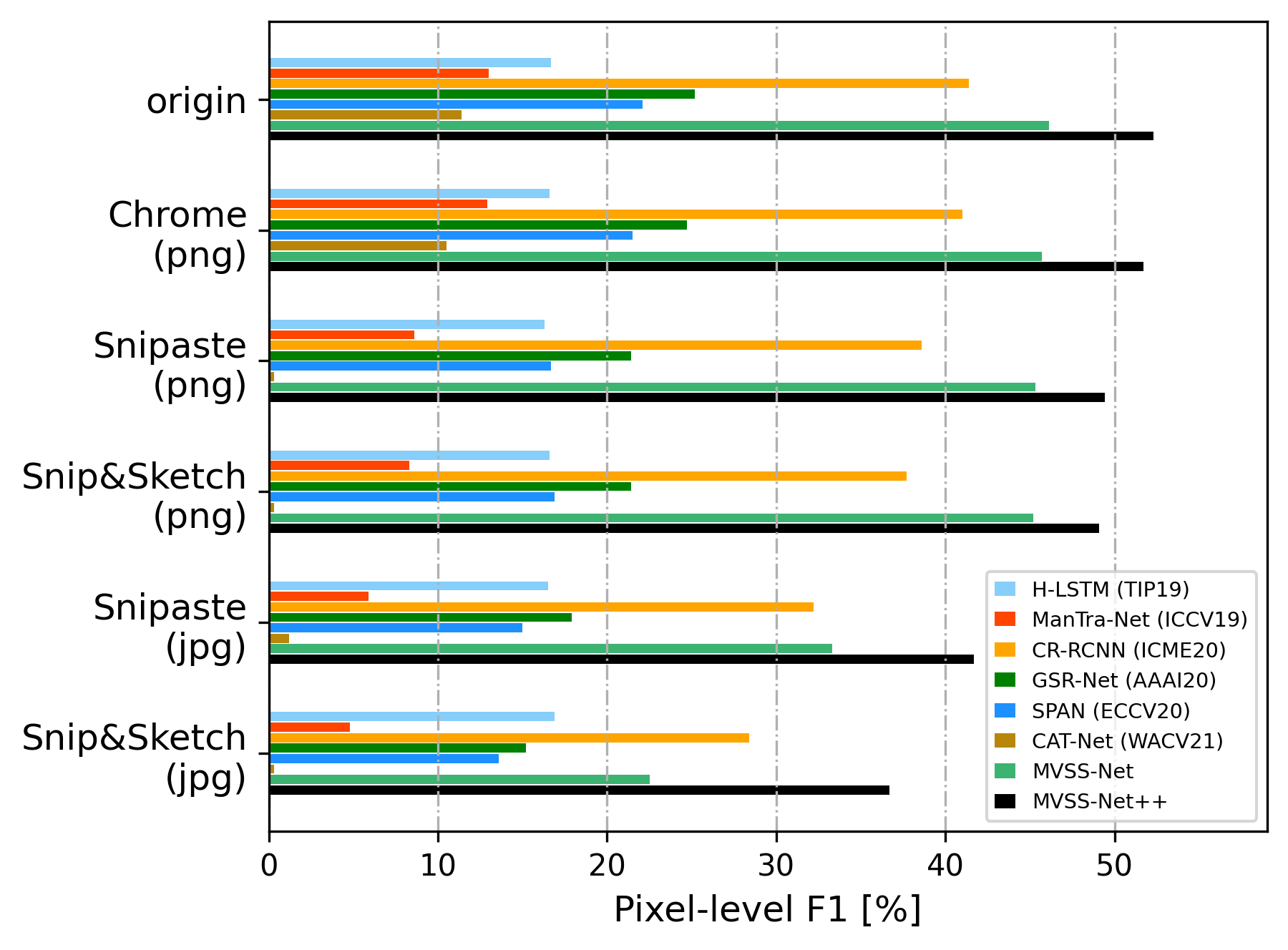}
    \caption{\textbf{Detection performance on images re-captured by three screenshot tools (Snip\&Sketch, Chrome and Snipaste) and saved in two different formats (jpg and png)}. Results are sorted by the performance of \modelplus ~in descending order.}
    \label{fig:recapture}
\end{figure}

Fig. \ref{fig:recapture} shows the performance of the individual models on the original test set and its variants. We draw two conclusions from the figure. First, concerning the two factors for image re-capturing, \ie screenshot tool and file format, the latter is more important. Second, while all models suffer from screenshot based image re-capturing, \modelplus ~remains the best.



%% file: table/table-fps.tex

 


\begin{table}[htbp!]
\caption{\textbf{Model inference speed}, tested on two NVIDA GPU cards respectively. Performance metric: Frames per second (FPS). Models are sorted in descending order in terms of their FPS on RTX2028ti, which is much cheaper than V100 and thus more affordable.}
\begin{center}
\renewcommand{\arraystretch}{1.1}
\footnotesize
\scalebox{1}{
\begin{tabular}{lrr}
\toprule
\textbf{Model} & \textbf{Tesla V100} & \textbf{RTX2080ti}\\\midrule
\model & 20.1 & \incolor{19.0} \\
\modelplus & 19.0 & 16.0\\
GSR-Net & \incolor{31.7} & 9.8 \\
SPAN & 8.4 & 8.1\\
H-LSTM & 6.5 & 5.4\\
CAT-Net & 5.4 & 4.1\\
C-RCNN & 2.8 & 2.2\\
ManTra-Net & 3.1 & 2.1\\
\bottomrule


 
\end{tabular}%
}
\end{center}

\label{table:fps}
\end{table}

%% file: exp-limitations.tex

\begin{figure}[htbp]
    \centering
    \includegraphics[width=\linewidth]{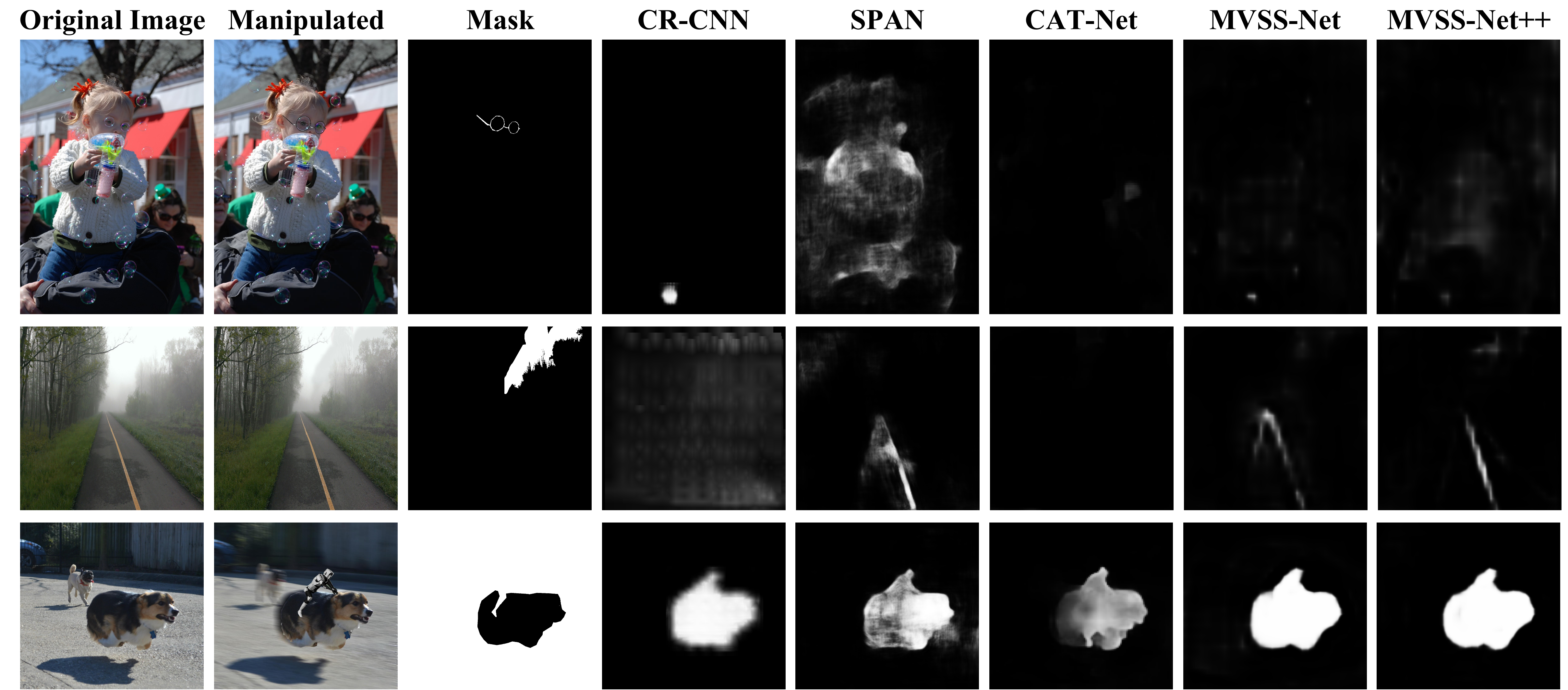}
    \caption{\textbf{Failure cases}. Data source: IMD.}
    \label{fig:fail}
\end{figure}

Given the challenging nature of the task, failures are inevitable, see Fig. \ref{fig:fail}. The first-row image was manipulated by darkening the frame of the spectacle the kid was wearing. Such manipulation traces appear to be too tiny to be revealed by the current models. The top-right corner of the second-row image was overlaid with certain translucent image patch, with the manipulated traces well melting into the misty scene. As for the last image, manipulation was performed by putting a knight on the back of the dog in the foreground, while blurring 
the background. All models capture the inconsistency between the foreground and the background, yet all fail to recognize that the background was actually manipulated.

%% file: conclusion.tex
For learning semantic-agnostic features, both noise and edge information are helpful, whilst the latter is better when used alone. For exploiting the edge information, our proposed edge-supervised branch (ESB) is more effective than the previously used feature concatenation. ESB steers the network to be more concentrated on tampered regions. Regarding the specificity of manipulation detection, we empirically show that the state-of-the-arts suffer from poor specificity. The inclusion of the image classification loss improves the specificity, yet at the cost of a clear performance drop for pixel-level manipulation detection. To avoid such a loss, multi-view feature learning has to be used together with multi-scale supervision. 
The resultant \modelplus~is a new state-of-the-art for image manipulation detection, outperforming the current methods in both within-dataset and cross-dataset scenarios. It also exhibits better robustness against JPEG compression, Gaussian blur and screenshot based image re-capturing. 


\lxr{With the initial success of MVSS-Net, we believe it will be  promising to design a more complex network that contains more components to cover other information (\eg compression artifacts) and other modalities (\eg associated text) for media forensics.}


%% file: appendix.tex
\lxr{\textbf{Additional measures}. Table \ref{table:app-metrix} shows accuracy and MCC scores of different models. The MVSS-Net series clearly outperform the baselines in terms of the well balanced MCC.}

\input{table/table-supp-acc}



\begin{figure}[htbp]
\begin{center}

\subfigure[A median filtering residual block (MFR)]{
\begin{minipage}[t]{0.95\linewidth}
\begin{center}
\includegraphics[width=0.55\columnwidth]{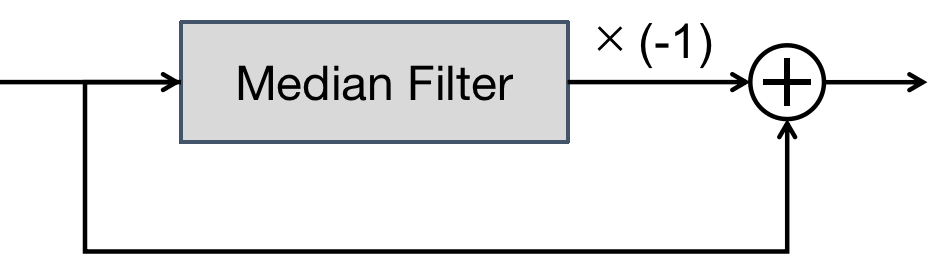}
\end{center}
\label{fig:mfr}
\end{minipage}%
}%

\subfigure[NSB with MFR]{
\begin{minipage}[t]{\linewidth}
\begin{center}
\includegraphics[width=\columnwidth]{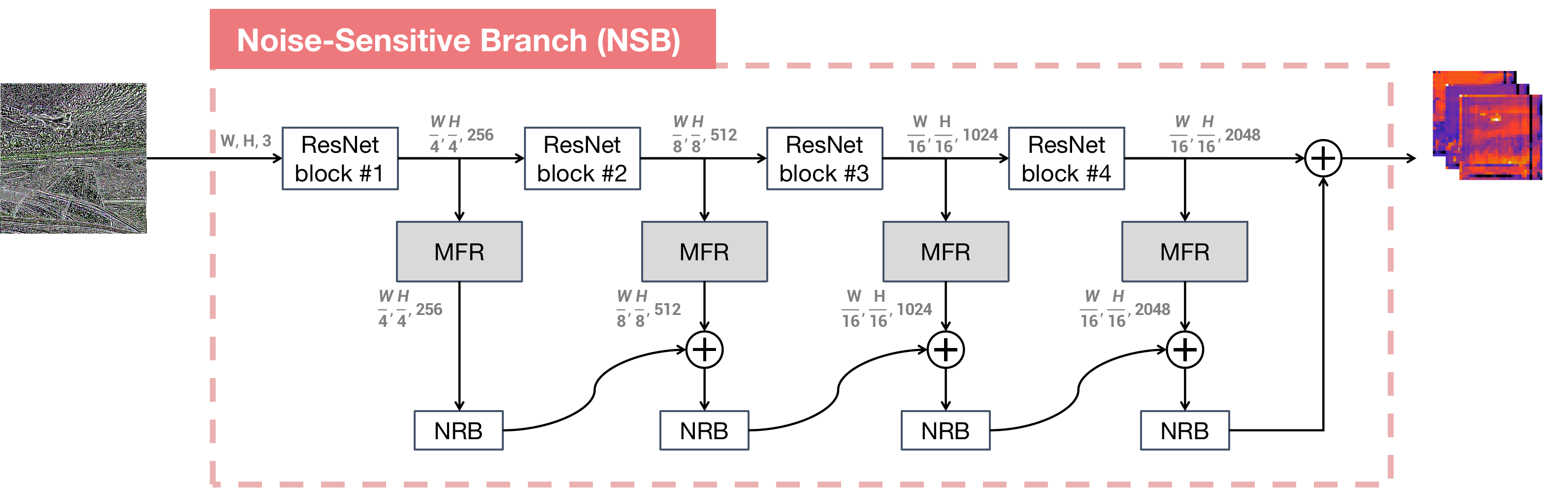}
\end{center}
\label{fig:mfr_model}
\end{minipage}%
}%

\end{center}
\caption{\lxr{\textbf{Diagrams of (a) non-trainable MFR  and (b) NSB with MFR}.}
}
\label{fig:nsb_mfr}
\end{figure}

\lxr{\textbf{NSB with MFR}. Fig. \ref{fig:nsb_mfr} shows how to add non-trainable MFR blocks to NSB, in a shallow-to-deep manner similar to ESB. NRB (noise residual block) is implemented in the same manner as ERB (edge residual block) in Fig. \ref{fig:erb}.}

%% file: table/table-supp-acc.tex
\begin{table}[htbp]
\caption{\lxr{\textbf{Detection performance measured by accuracy and MCC}.}}
\renewcommand{\arraystretch}{1}
\begin{center}

\scalebox{0.77}{
\begin{tabular}{@{}lrrrrrrr@{}}
\toprule
{\textbf{Method}} & \textbf{NIST} & \textbf{Columbia} & \textbf{CASIAv1+}  & \textbf{COVER}  & \textbf{DEF-12k}  & \textbf{IMD} & \textbf{MEAN}\\\midrule


\multicolumn{6}{@{}l}{\textbf{\emph{Pixel-level accuracy (\%):}}}\\
H-LSTM &92.8 & 69.2 & 90.3 & 87.4 & 94.3 & 90.9 & 87.5\\  
ManrTra-Net & 92.5 & 74.6 & 88.2 & 90.2 & 96.9 & 92.3 & 89.1\\  
C-RCNN & \textbf{92.9} & 77.2 & 80.5 & 89.1 & 96.8 & 91.4 & 88.0\\  
GSR-Net & 88.4 & 80.3 & 87.9 & 84.9 & 93.7 & 90.2 & 87.6\\  
SPAN &88.8 & 77.3 & 91.6 & 88.2 & 95.0 & 91.0 & 88.7\\  
CAT-Net& 92.1 & \textbf{82.0} & 91.9 & 90.0 & 96.8 & 92.5 & \textbf{90.9}\\ 
FCN &92.4 & 70.8 & 93.6 & 88.0 & 96.8 & \textbf{92.4} & 89.0\\  
\model &90.1 & 77.6 & \textbf{94.0} & 91.1 & \textbf{97.0} & 91.1 & 90.2 \\
\modelplus &90.5 & 66.0 & 93.1 & \textbf{91.4} & 96.8 & 91.0 & 88.1 \\ \midrule

\multicolumn{7}{@{}l}{\textbf{\emph{Image-level accuracy  (\%):}}}\\
H-LSTM& 92.8&	50.1 &	53.3 &	50.0 &	50.0 &	\textbf{82.9} &	63.2 \\
ManrTra-Net&92.5&	49.6 &	53.5 &	50.0 &	50.0 &	\textbf{82.9} &	63.1\\ 
C-RCNN&\textbf{92.9}&	60.1 &	56.2 &	51.9 &	52.1 &	79.1 &	65.4 \\
GSR-Net&88.4&	50.1 &	53.2 &	50.0 &	45.8 &	\textbf{82.9} &	61.7 \\
SPAN&88.8&	49.6 &	53.5 &	50.0 &	50.0 &	\textbf{82.9} &	62.5 \\
CAT-Net&92.1&	91.7 &	55.6 &	54.0 &	53.4 &	36.7 &	63.9\\
FCN&92.4&	63.3 &	68.8 &	50.0 &	52.5 &	72.8 &	66.6 \\
\model&90.1&	83.6 &	\textbf{78.8} &	54.0 &	\textbf{54.3} &	79.6 &	\textbf{73.4}\\ 
\modelplus&90.5&	\textbf{93.1} &	74.4 &	\textbf{68.5} &	52.0 &	60.2 &	73.1 \\\midrule
 
\multicolumn{7}{@{}l}{\textbf{\emph{Pixel-level MCC [-1, 1]:}}}\\
H-LSTM & \textbf{0.351} & 0.124 & 0.138 & 0.131 & 0.046 & 0.182 & 0.162\\ 
ManrTra-Net &0.000 & 0.365 & 0.092 & 0.313 & \textbf{0.175} & 0.194 & 0.190\\  
C-RCNN &0.232 & 0.408 & 0.380 & 0.273 & 0.140 & 0.254 & 0.281  \\
GSR-Net &0.257 & 0.518 & 0.178 & 0.228 & 0.083 & 0.224 & 0.248  \\
SPAN &0.203 & 0.444 & 0.190 & 0.164 & 0.039 & 0.161 & 0.200  \\
CAT-Net& 0.175 & 0.518 & 0.138 & 0.127 & 0.048 & 0.058 & 0.177 \\
FCN& 0.151 & 0.194 & 0.425 & 0.154 & 0.113&  0.212 & 0.208  \\
\model& 0.279 & 0.492 & 0.447&  0.437 & 0.099 & 0.256 & 0.335 \\ 
\modelplus& 0.289 & \textbf{0.545} & \textbf{0.503} & \textbf{0.464} & 0.097&  \textbf{0.265} & \textbf{0.361}  \\\midrule
 
\multicolumn{7}{@{}l}{\textbf{\emph{Image-level MCC [-1, 1]:}}}\\
H-LSTM & --&	0.074 &	-0.039 &	0.000 &	-0.009 &	-0.009 &	0.003\\ 
ManTra-Net &--	&0.000 &	0.000 &	0.000 &	0.000 &	0.000 &	0.000\\ 
C-RCNN &--	&0.295 &	0.114 &	0.084 &	0.048 &	0.073 &	0.123\\ 
GSR-Net &--&	0.074 &	-0.053 &	0.000 &	-0.208& 0.000 &	-0.037 \\
SPAN &--&	0.000 &	0.000 &	0.000 &	0.000 &	0.000 &	0.000 \\
CAT-Net&--&	0.838 &	0.216 &	0.094 &	0.072 &	0.078 &	0.259 \\
FCN&--&	0.349 &	0.372 &	0.000 &	0.053 &	0.001 &	0.155 \\
\model&--&	0.710 &	\textbf{0.637} &	0.133 &	\textbf{0.102} &	0.163 &	0.349 \\
\modelplus&--&	\textbf{0.865} &	0.569 &	\textbf{0.370} &	0.041 &	\textbf{0.174} &	\textbf{0.404} \\

\bottomrule
\end{tabular}
}
\end{center}

\label{table:app-metrix}
\end{table}

%% file: acknowledgment.tex

\section*{Acknowledgments}
This research was supported by NSFC (62172420, U1703261), BJNSF (4202033), the Fundamental Research Funds for the Central Universities and the Research Funds of Renmin University of China (No. 18XNLG19), and Public Computing Cloud, Renmin University of China. The authors thank the anonymous reviewers for their  insightful feedback.

%% file: biography.tex



\begin{IEEEbiography}[{\includegraphics[width=1in,height=1.25in,clip,keepaspectratio]{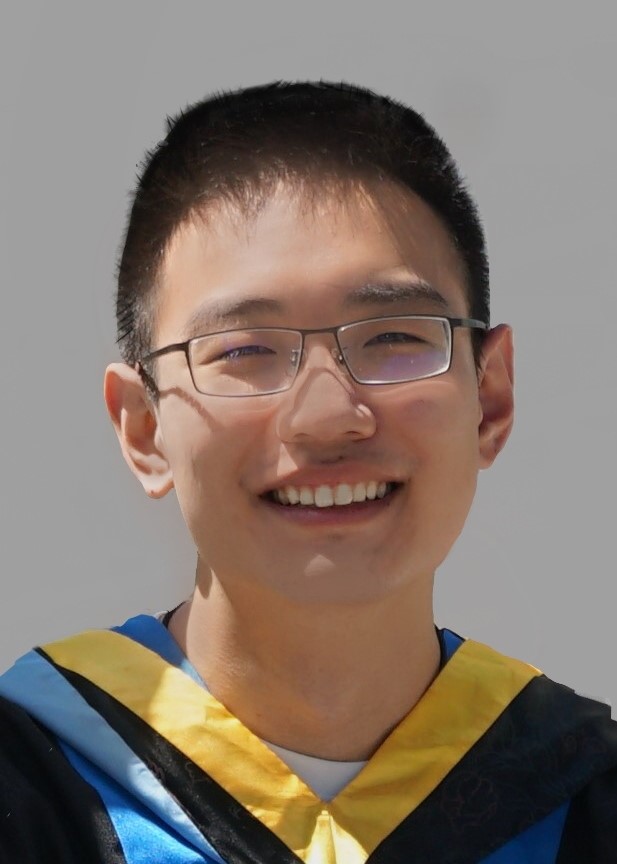}}]
{Chengbo Dong} received his B.S. degree in automation from Beihang University, Beijing, China in 2020. He is currently a master student at the AIMC Lab, School of Information, Renmin University of China, pursuing his master degree on multimedia forensics.
\end{IEEEbiography}

\begin{IEEEbiography}[{\includegraphics[width=1in,height=1.25in,clip,keepaspectratio]{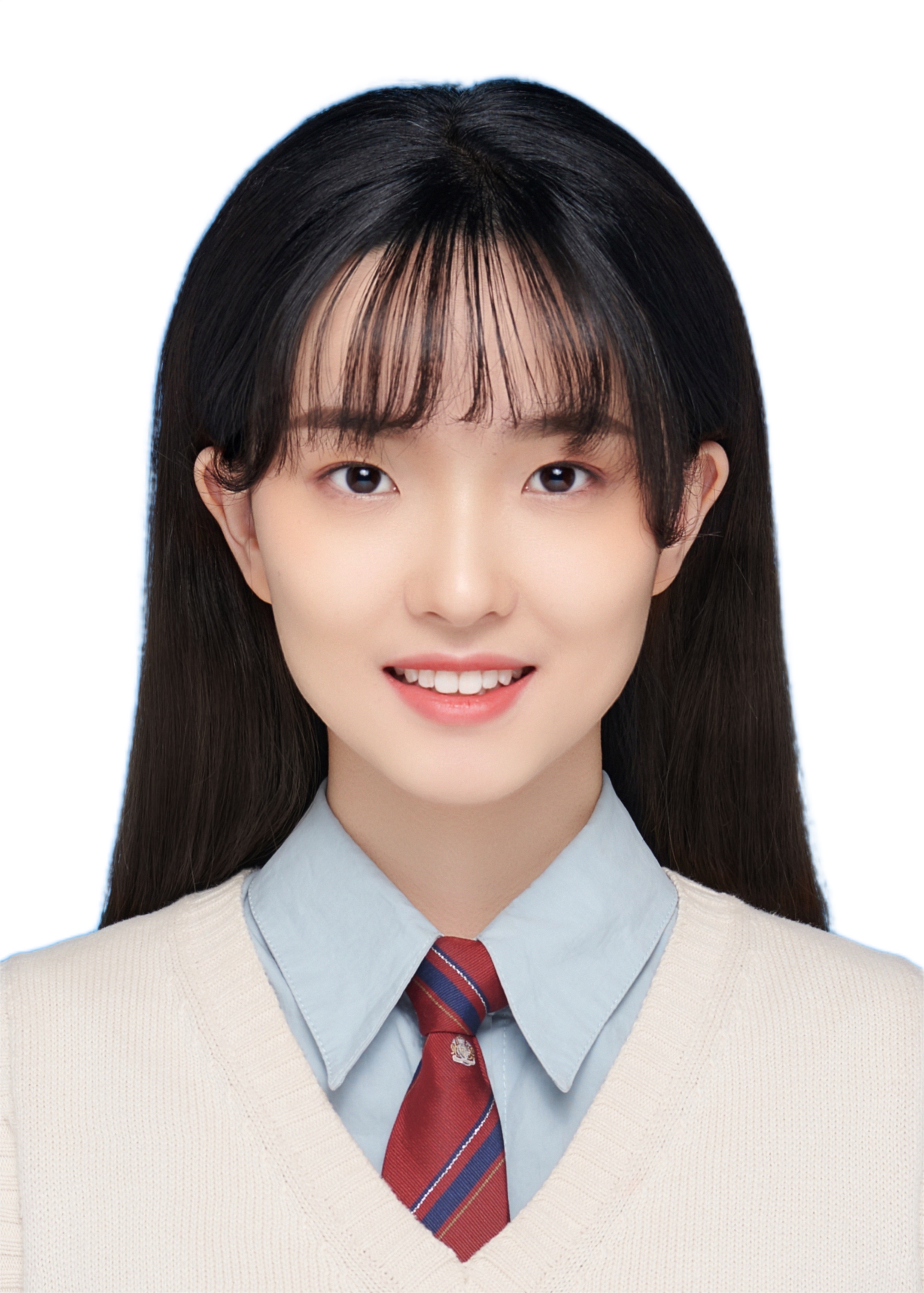}}]
{Xinru Chen} received her B.S. degree in Computer Science from the College of Information, Renmin University of China, Beijing, China in 2020. She is currently a master student at the AIMC Lab, School of Information, Renmin University of China, pursuing her master degree on multimedia forensics.
\end{IEEEbiography}

\begin{IEEEbiography}[{\includegraphics[width=1in,height=1.25in,clip,keepaspectratio]{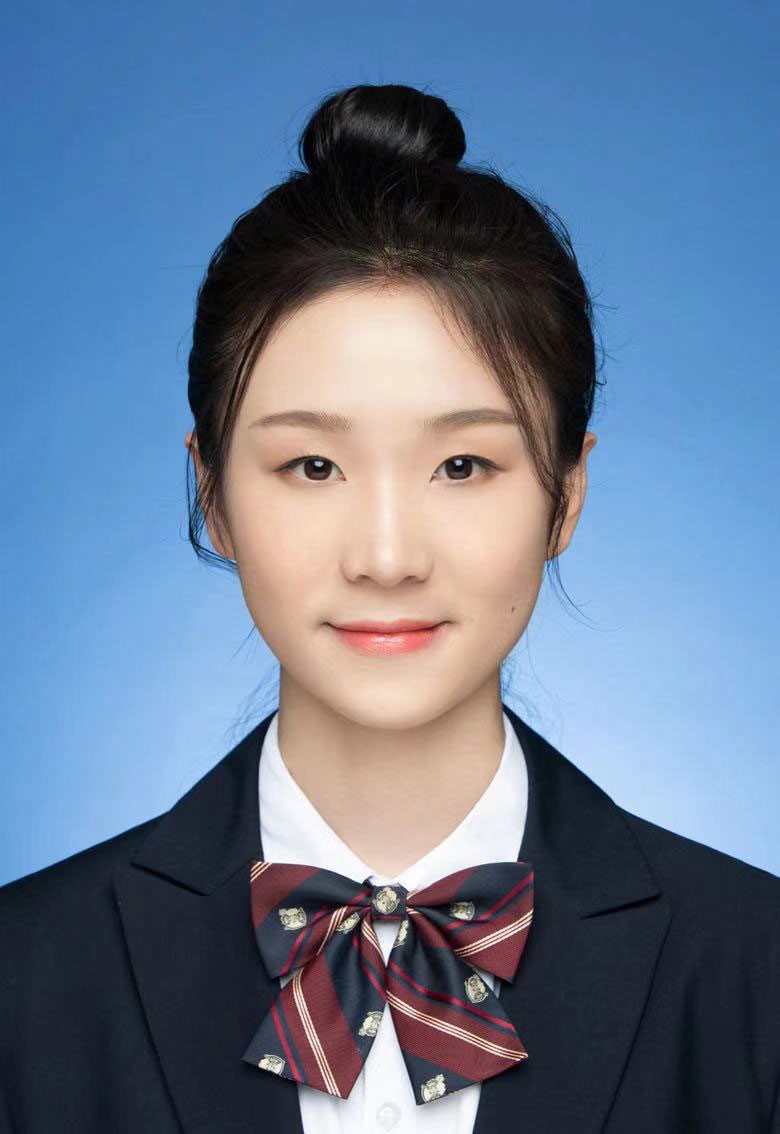}}]
{Ruohan Hu} is an undergraduate at the College of Information, Beijing Forestry University. She is currently a research intern at the AIMC Lab, School of Information, Renmin University of China.
\end{IEEEbiography}

\begin{IEEEbiography}[{\includegraphics[width=1in,height=1.25in,clip,keepaspectratio]{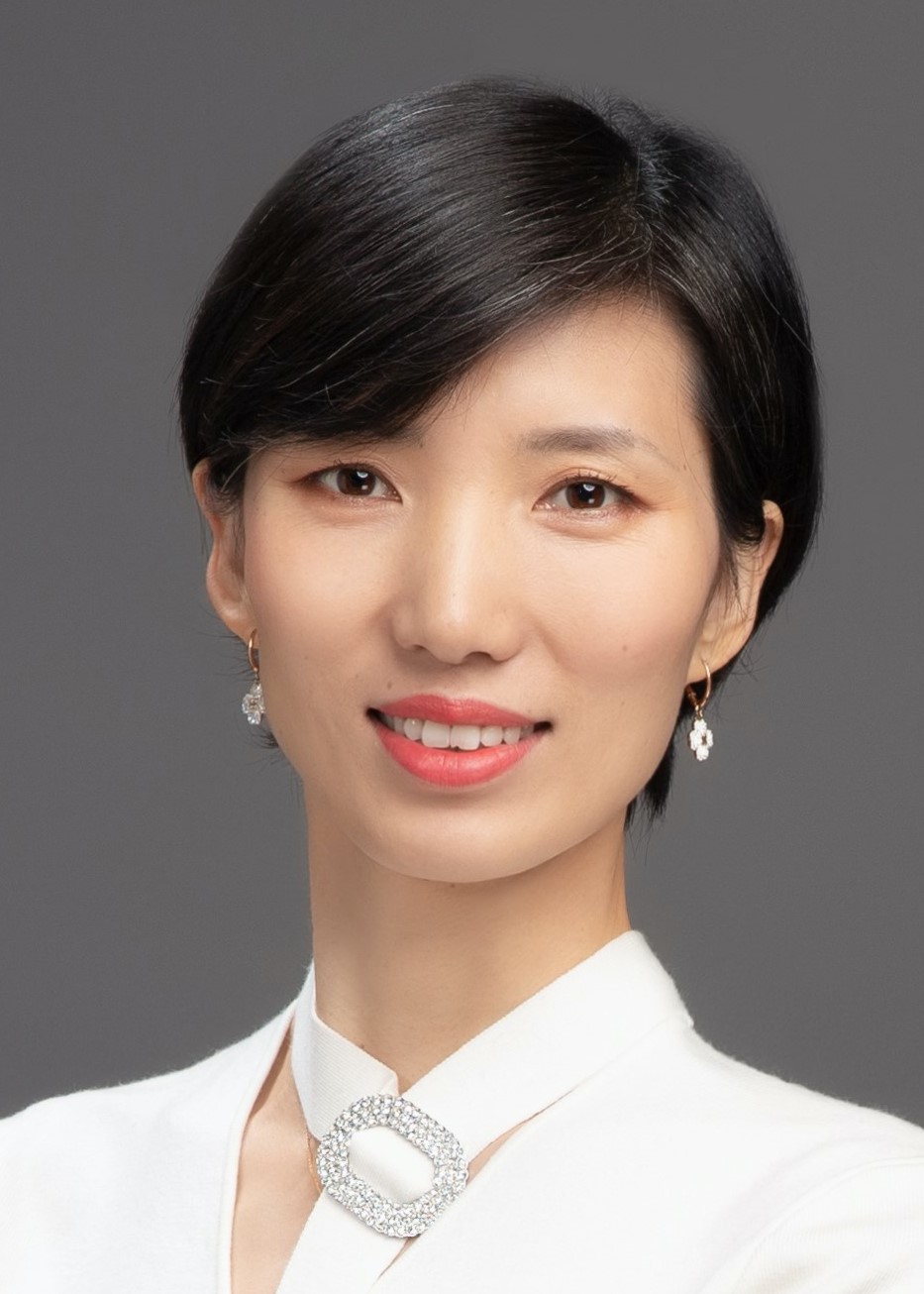}}]
{Juan Cao} received her Ph.D. degree from the Institute of Computing Technology,
Chinese Academy of Sciences, Beijing, China, in 2008. She is currently a Full Professor with the same institute. Her research interests include multimedia content analysis, fake news detection and forgery detection.
\end{IEEEbiography}

\begin{IEEEbiography}[{\includegraphics[width=1in,height=1.25in,clip,keepaspectratio]{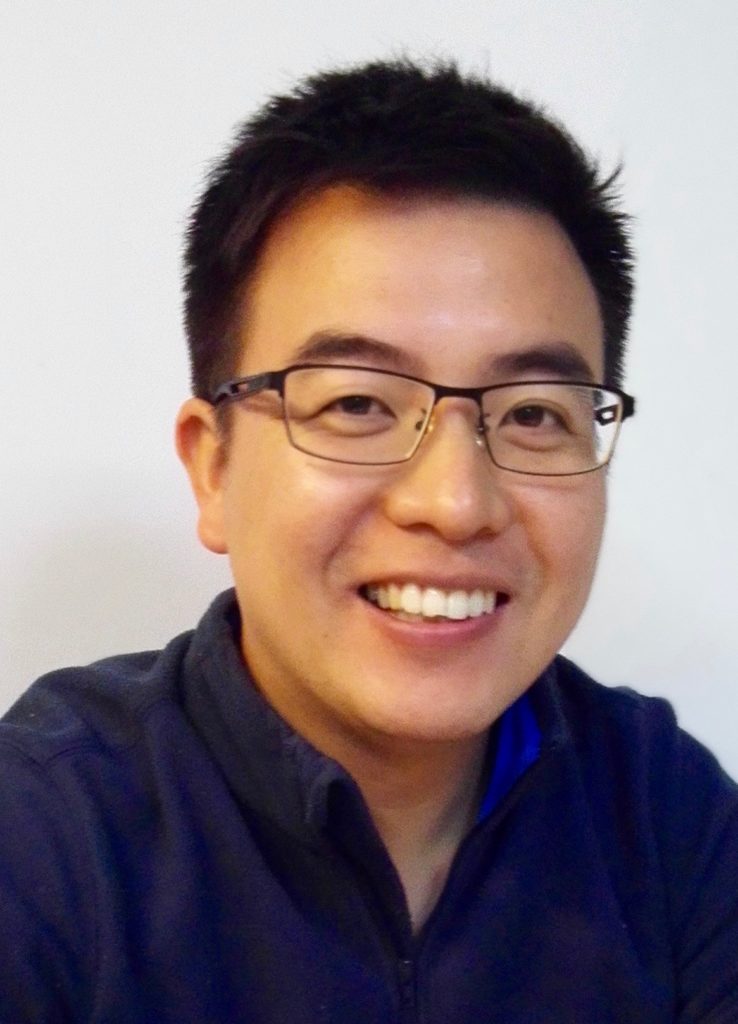}}]
{Xirong Li} received the B.S. and M.E. degrees from Tsinghua University, Beijing, China, in 2005 and 2007, respectively, and the Ph.D. degree from the University of Amsterdam, Amsterdam, The Netherlands, in 2012, all in computer science. He is currently an Associate Professor with the Key Lab of Data Engineering and Knowledge Engineering, Renmin University of China, Beijing, China. He leads the AIMC Lab at RUC. His research focuses on multimedia intelligence.
Dr. Li was recipient of the ACMMM 2016 Grand Challenge Award, the ACM SIGMM Best Ph.D. Thesis Award 2013, the IEEE TRANSACTIONS ON MULTIMEDIA Prize Paper Award 2012, and the Best Paper Award of ACM CIVR 2010. He served as program co-chair of the Multimedia Modeling 2021 conference and is serving as associate editor of ACM TOMM and the Multimedia Systems Journal.
\end{IEEEbiography}